%% file: main.tex
\documentclass[lettersize,journal,twoside]{IEEEtran}
\usepackage[scaled=0.95]{newtxtext}
\usepackage{amsmath}
\usepackage{amssymb}
\usepackage{algorithmic}
\usepackage{algorithm}
\usepackage{array}
\usepackage[caption=false,font=normalsize,labelfont=sf,textfont=sf]{subfig}
\usepackage{textcomp}
\usepackage{stfloats}
\usepackage{url}
\usepackage{graphicx}
\usepackage{cite}
\usepackage{bm}
\usepackage[hidelinks]{hyperref}
\usepackage{orcidlink}
\usepackage{multirow}
\usepackage{booktabs}
\usepackage{tabularx}
\usepackage{makecell}
\usepackage{siunitx}
\IEEEsetsidemargin{c}{-2pt}
\makeatletter
\def\@IEEEheaderstyle{\normalfont\fontsize{7.4pt}{8pt}\selectfont}
\makeatother
\begin{document}
\bstctlcite{IEEEtran:BSTcontrol}

\title{{\fontsize{24.8pt}{28pt}\selectfont
\scalebox{1.02}[1]{CC-4DGS: Computational Deformation and}\\[-0.7pt]
\scalebox{1.027}[1]{Point-Cloud Compression for Storage-Efficient}\\
\scalebox{1.02}[1]{Dynamic Gaussian Splatting}}}

\author{Kyungdae Park\orcidlink{0009-0009-1060-3730},~\textit{Graduate Student Member,~IEEE},%
        ~and~Chae Eun Rhee\orcidlink{0000-0002-7851-1703},~\textit{Senior Member,~IEEE}%
\thanks{Received 11 December 2025; revised 21 July 2026; accepted 3 August\linebreak 2026. This work was supported in part by the National Research Foundation of Korea (NRF) grant funded by the Korea government (MSIT) under Grant RS-2025-00562356, in part by the Institute of Information \& Communications Technology Planning \& Evaluation (IITP) through the Information Technology Research Center (ITRC) grant funded by the Korea government (MSIT) under Grant RS-2021-II212052, and in part by the Institute of Information \& Communications Technology Planning \& Evaluation (IITP) through the Artificial Intelligence Semiconductor Support Program to Nurture the Best Talents, funded by the Korea government (MSIT), under Grant IITP-2026-RS-2023-00253914. Recommended for acceptance by J. Hou. \emph{(Corresponding author: Chae Eun\linebreak Rhee.)}}
\thanks{Kyungdae Park is with the Department of Artificial Intelligence Semiconductor Engineering, Hanyang University, Seoul 04763, South Korea (e-mail: kdpark@hanyang.ac.kr).}
\thanks{Chae Eun Rhee is with the Department of Electronic Engineering, Hanyang University, Seoul 04763, South Korea (e-mail: crhee@hanyang.ac.kr).}
\thanks{The source code is publicly available at \url{https://github.com/KyungdaePark/CC-4DGS}.}
\thanks{Digital Object Identifier \href{https://doi.org/10.1109/TVCG.2026.3721136}{10.1109/TVCG.2026.3721136}}
}

\markboth{IEEE Transactions on Visualization and Computer Graphics, Vol.~0, No.~0, 2026}%
{Park and Rhee: CC-4DGS: Storage-Efficient Dynamic Gaussian Splatting}

\makeatletter
\def\@IEEEpubidpullup{2.3\baselineskip}
\makeatother
\IEEEpubid{%
\raisebox{4.7pt}[0pt][0pt]{\scalebox{1.09}[1]{\parbox{\textwidth}{\centering\fontsize{7.7pt}{9.3pt}\selectfont
1077-2626 \copyright~2026 IEEE. All rights reserved, including rights for text and data mining, and training of artificial intelligence and similar technologies.\\
Personal use is permitted, but republication/redistribution requires IEEE permission. See \href{https://www.ieee.org/publications/rights/index.html}{https://www.ieee.org/publications/rights/index.html} for more information.}}}}

\IEEEaftertitletext{\vspace{-5.33pt}}
\maketitle

\begin{abstract}
\bfseries
Dynamic four-dimensional (4D) Gaussian Splatting\linebreak has emerged as a powerful explicit representation for high-quality view synthesis, yet existing methods still require tens to hundreds of megabytes per scene due to their heavy reliance on large multi-resolution hash tables and high-dimensional Gaussian attributes. This paper presents CC-4DGS, a storage-efficient and scalable framework that rethinks both deformation modeling and canonical attribute storage. First, we introduce a computational deformation field (CDF) that replaces large multi-resolution learnable hash tables with deterministic dense hash encoding and compact neural decoders, enabling on-the-fly synthesis of deformation features while reducing deformation storage to only 1--3 MB per scene. Second, we propose a compression of canonical point-cloud attributes (CCA) pipeline that compresses high-dimensional spherical harmonic appearance terms and auxiliary Gaussian attributes via conditional autoencoding, selective quantization, and residual codebooks, achieving 3--5$\times$ point-cloud reduction with negligible quality loss. Together, these components yield a unified representation that preserves real-time rendering performance while reducing total storage to 20--30 MB. Extensive experiments across the N3DV and Technicolor Light Field datasets demonstrate that CC-4DGS achieves reconstruction accuracy comparable to state-of-the-art methods such as Swift4D, while offering significantly improved storage efficiency and favorable runtime--memory trade-offs.
\end{abstract}
\vspace{2pt}

\vspace{-6.8pt}

\section{Introduction}
\vspace{2.03pt}
\noindent
\input{Sections/Introduction}

\section{Related Works}
\noindent
\input{Sections/Related_works}

\section{Preliminaries}
\noindent
\input{Sections/Preliminaries}

\section{Proposed CC-4DGS Framework}
\noindent
\input{Sections/Method}

\section{Performance Evaluation}
\noindent
\input{Sections/Performance_Evaluation}

\section{Ablation Study}
\noindent
\input{Sections/Ablation_Study}

\section{Limitations and Future Directions}
\label{sec:limitations}
\input{Sections/Limitations}

\section{Conclusion}
In this work, we introduced CC-4DGS, a storage-efficient framework for dynamic Gaussian splatting that jointly addresses the deformation and point-cloud storage bottlenecks of prior 4DGS systems. By replacing multi-resolution hash tables with the CDF and compressing canonical Gaussian attributes through the CCA pipeline, our method achieves substantial reductions in both deformation and appearance storage while maintaining high visual fidelity and real-time rendering performance. Experiments on the N3DV and Technicolor datasets demonstrate that CC-4DGS offers a significantly more favorable trade-off among storage, reconstruction quality, and runtime efficiency than existing approaches. Future work will explore more adaptive hashing and neural decoding strategies, as well as extensions to longer and more complex dynamic sequences and real-time streaming scenarios.

\IEEEtriggeratref{37}
\bibliographystyle{IEEEtran}
\bibliography{reference}

\makeatletter
\def\@IEEEBIOhangwidth{1.16in}
\makeatother
\vspace{16pt}
\begin{IEEEbiography}[{\includegraphics[width=1in,height=1.25in,clip,keepaspectratio]{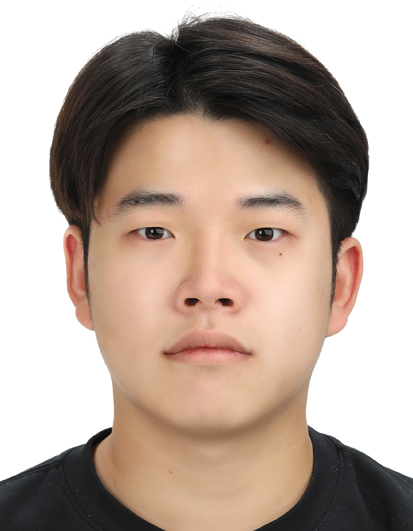}}]{Kyungdae Park} (Graduate Student Member, IEEE) received the BS degree in information and communication engineering from Inha University, Incheon, South Korea, in 2024. He is currently working toward the integrated MS--PhD degree with the Department of Artificial Intelligence Semiconductor Engineering, Hanyang University, Seoul, South Korea. His research interests include neural rendering, light field video processing, and dynamic scene reconstruction.
\end{IEEEbiography}
\vspace{24.67pt}
\begin{IEEEbiography}[{\includegraphics[width=1in,height=1.25in,clip,keepaspectratio]{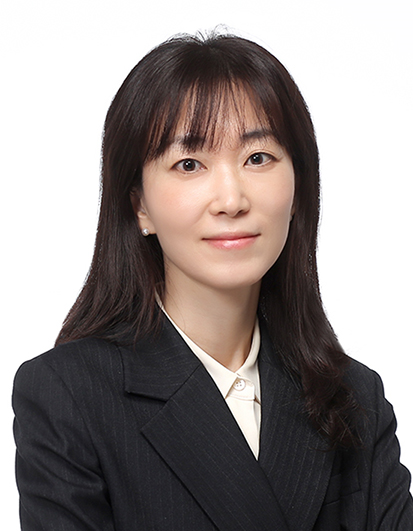}}]{Chae Eun Rhee} (Senior Member, IEEE) received the BS, MS, and PhD degrees in electrical engineering and computer science from Seoul National University, Seoul, South Korea, in 2000, 2002, and 2011, respectively. From 2002 to 2005, she was an engineer with Digital TV Development Group, Samsung Electronics Company, Ltd., Suwon City, Korea, focusing on bus architecture and MPEG decoder development. From 2013 to 2023, she was a faculty member with the Department of Electrical and Computer Engineering, Inha University. In 2024, she joined the Department of Electronic Engineering, Hanyang University, Korea, where she is currently a professor. Her research interests include video coding algorithms and architectures, immersive video systems, and hardware design for AI applications.
\end{IEEEbiography}

\vfill

\end{document}

%% file: Sections/Introduction.tex
\IEEEPARstart{T}{hree}-dimensional (3D) scene representation has be-\linebreak come a fundamental component in various fields such\newpage\noindent as robotics, autonomous driving, and virtual/augmented reality.\linebreak Recently, there has been a growing demand for four-dimensional (4D) video data, which captures not only static 3D scenes but also dynamic changes over time—beyond the limitations of conventional two-dimensional (2D) imagery. 4D video enables natural and consistent scene rendering under varying viewpoints and environmental conditions, significantly enhancing immersion and system responsiveness. However, such high-dimensional data inherently requires substantial storage, posing serious challenges for transmission and long-term archiving, especially under limited network bandwidth and storage capacity. In mobile and edge computing environments in particular, real-time processing or streaming of high-resolution video remains highly constrained. Therefore, developing technologies for the efficient compression, storage, and rendering of 3D and 4D data is critical, and the importance of such research continues to grow.

Contemporary neural network-based 3D video techniques\linebreak are broadly categorized into implicit and explicit representations. Implicit approaches such as Neural Radiance Fields (NeRF)~\cite{nerf} offer high visual fidelity but require thousands of multi-layer perceptron (MLP) evaluations per ray, making them impractical for real-time applications. In contrast, explicit methods including 3D Gaussian Splatting (3DGS)~\cite{3dgs}, Plenoxels~\cite{plenoxels}, and Instant-NGP~\cite{ingp} enable fast rendering by storing large-scale voxel grids or Gaussian primitives. When extended to 4D video, however, explicit representations incur substantial memory overhead, as seen in Grid4D~\cite{grid4d}, 4DGS~\cite{4dgs_wu}, and SaRO-GS~\cite{sarogs}, which often require hundreds of megabytes to several gigabytes due to per-frame parameters or massive hash tables along the temporal axis. To mitigate these costs, recent techniques such as Scaffold-GS~\cite{scaffoldgs} and MoDec-GS~\cite{modecgs} reduce redundancy through parameter sharing, pruning, and entropy-based regularization, bringing storage down to several tens of megabytes; yet they still rely on millions of explicit parameters, limiting scalability for network streaming, mobile\linebreak and edge deployment, and cloud storage. More recent efforts—including Motion-Aware 3DGS~\cite{guo2024motion}, FRPGS~\cite{Li2025FRPGS}, and StreetSurfGS~\cite{Cui2025StreetSurfGS}—further improve efficiency by removing Gaussian redundancy, and Swift4D~\cite{Swift4d} demonstrates competitive reconstruction quality on challenging dynamic scenes. Nevertheless, even state-of-the-art dynamic 4DGS pipelines remain dependent on explicit Gaussian structures, large hash tables, and high-dimensional per-Gaussian features, keeping the per-scene \mbox{footprint} in the tens-of-megabytes range. While compression\newpage techniques can alleviate some redundancy, it remains insufficient to capture the diversity and temporal complexity of dynamic scenes, indicating the need for a more fundamental rethinking of data structures and representation strategies.\IEEEpubidadjcol

We target dynamic 4DGS methods that use a canonical Gaussian set with a learned deformation field, such as 4DGS~\cite{4dgs_wu}, Swift4D~\cite{Swift4d}, and Grid4D~\cite{grid4d}, rather than native 4D primitive representations such as 4D-GS~\cite{4dgs_yang}. To overcome the fundamental storage bottlenecks of 4DGS, we introduce a unified two-fold strategy that rethinks both the deformation and appearance storage of Gaussian representations. First, we replace the large multi-resolution learnable hash tables used by prior dynamic 4DGS methods with a computational deformation field (CDF), in which deformation features are computed on-the-fly by a compact decoder rather than retrieved from a stored hash table. Concretely, the CDF is implemented as a compact multi-MLP module that directly predicts 4D motion from dense hash-encoded coordinates, eliminating the dominant source of storage overhead in prior work such as Grid4D and Swift4D.  Our computational deformation module remains highly expressive while reducing the deformation footprint to only 1–3 MB per scene.
Second, we introduce a learned compression of canonical point-cloud attributes (CCA) pipeline that targets the canonical Gaussian attributes—particularly the high-dimensional spherical harmonic (SH) appearance terms, which account for the majority of storage in existing dynamic 3DGS and 4DGS systems. Through conditional autoencoding, quantization, and residual codebook modeling, our method compresses SH components and auxiliary attributes with negligible quality loss, reducing the point-cloud storage by 3–5× compared to uncompressed canonical representations.
Together, CDF and CCA form CC-4DGS, a unified framework that substantially reduces both deformation and appearance storage without sacrificing visual fidelity or real-time rendering speed. CC-4DGS achieves competitive reconstruction accuracy compared to recent high-quality methods such as Swift4D, while lowering Deform-module storage to $1$--$3$~MB and total storage to 20–30 MB per scene.

%% file: Sections/Related_works.tex
\subsection{Neural Representations for Static and Dynamic Scenes}
Neural radiance fields (NeRF)~\cite{nerf} model view-dependent color and density as an implicit MLP, achieving high-fidelity novel view synthesis but requiring thousands of network evaluations per ray. To improve efficiency, explicit voxel-based representations such as Plenoxels~\cite{plenoxels} and NSVF~\cite{nsvf} replace most MLP queries with fast grid interpolation. Instant-NGP~\cite{ingp} further introduces multi-resolution hash grids that adapt spatial resolution to scene complexity, enabling real-time training and rendering. Building on this line of work, 3D Gaussian Splatting (3DGS)~\cite{3dgs} represents scenes with explicit Gaussian primitives and rasterization-based splatting, offering interpretable parameters and high frame rates.

Extending static representations to dynamic scenes requires modeling temporal coherence in addition to spatial structure. Early dynamic NeRF methods augment the radiance field with time and directly learn a 4D plenoptic function~\cite{dnerf,nerfies}, while later works such as HexPlane~\cite{hexplane} and K-Planes~\cite{kplanes} factor the 4D domain into low-rank or multi-plane grids. Canonical-space approaches~\cite{dnerf,nerfies} instead learn a deformation field from a static template to observed frames, improving temporal consistency but still incurring significant MLP cost. However, as scene complexity and motion diversity increase, implicit formulations often struggle to balance accuracy, computational overhead, and storage, motivating a shift toward more explicit and modular representations. Our work follows the explicit Gaussian paradigm rather than implicit NeRF, and focuses on reducing the storage footprint of dynamic 3DGS-style representations.

\subsection{Dynamic Gaussian Splatting and Storage-Efficient Extensions}

Dynamic Gaussian Splatting methods aim to exploit the explicit and compact nature of Gaussian primitives while modeling time-varying scenes. A widely adopted formulation defines Gaussians in a canonical space and learns a deformation field that maps them to observed states over time. Deformable 3D Gaussians~\cite{deformable3dgs}, 4DGS~\cite{4dgs_wu}, and GauFRe~\cite{gaufre} employ multi-resolution grids or compact MLPs to parameterize such deformations, while E-D3DGS~\cite{e-d3dgs} introduces per-Gaussian latent embeddings to enrich the deformation capacity beyond coordinate-based fields. Grid4D~\cite{grid4d} and Swift4D~\cite{Swift4d} further adopt Instant-NGP-style multi-resolution hash grids to predict canonical-to-observed offsets, achieving strong reconstruction quality at the cost of large hash tables and high-dimensional per-Gaussian features. Other works bypass a canonical template entirely by placing Gaussians directly in the 4D spatio-temporal domain $(x,y,z,t)$, as in 4D-GS~\cite{4dgs_yang}, Ex4DGS~\cite{ex4dgs}, SaRO-GS~\cite{sarogs}, and SpaceTime-GS~\cite{spacetimegs}. While this fully explicit formulation captures complex nonrigid motion, it often requires a large number of primitives, leading to model sizes that commonly reach hundreds of megabytes per scene. Recent monocular-input variants such as MoDGS~\cite{modgs} and Decoupling Dynamic Monocular Videos~\cite{decoupling-dynamic-monoc} extend dynamic Gaussian reconstruction to single-camera capture, targeting a different input setting than the multi-view dynamic 4DGS we address.

A complementary line of research focuses on reducing the storage demands inherent to dynamic Gaussian splatting. Scaffold-GS and MoDec-GS~\cite{scaffoldgs,modecgs} employ anchor-based feature sharing and temporally adaptive scaffolds to shrink parameter counts, though sizable feature tables or dense per-point attributes often remain. Additional efforts investigate more compact or streamable 4D Gaussian representations: 4DGC~\cite{4dgc} combines differentiable quantization with a lightweight implicit entropy model, while MEGA~\cite{mega} reduces memory through a compact color representation and entropy-constrained deformation, and 4DGV~\cite{4dgv} leverages an H.265-based motion-layering pipeline to compress deformation offsets. In an adjacent data domain, DPC-as-2D-Videos~\cite{dpc-2d-videos} re-represents dynamic point cloud sequences as 2D video for compression, targeting point-cloud rather than Gaussian dynamic data. GIFStream~\cite{gifstream} compresses deformation-field 4DGS through a feature-stream representation tailored to immersive video, and Light4GS~\cite{light4gs} further reduces 4DGS footprint via a context-model entropy coder. Dynamics-Aware Gaussian Splatting Streaming~\cite{dynamics_aware_streaming} extends this streaming direction with on-the-fly 4D reconstruction guided by motion-aware decomposition. DASH~\cite{dash} explores hash-based acceleration through a 4D hash encoding combined with dynamic–static decomposition for improved real-time rendering. Collectively, these approaches highlight the central trade-offs among deformation modeling, storage overhead, and reconstruction fidelity, underscoring the need for increasingly compact and scalable representations for dynamic 4D scenes.
Concurrently, Fov-GS~\cite{fov-gs} and FUGS~\cite{fugs} enrich dynamic 3D Gaussian splatting with foveated rendering and frequency-aware uncertainty modeling, improving real-time rendering performance and robustness to complex motion, yet they largely retain comparable parameter counts. Compression efforts targeting static 3DGS, such as CompGS~\cite{compgs} and EAGLES~\cite{eagles}, reduce per-Gaussian attribute storage through vector quantization or compact attribute encodings, but do not address deformation-field storage; their attribute-level compression strategies complement our canonical-side CCA pipeline rather than serving as direct dynamic-4DGS baselines. In contrast, we redesign the deformation and point-cloud representations to reduce storage footprint and improve the rate–distortion trade-off.

\subsection{Hash-Based Embeddings and Learnable Hash-Table-Free Feature Learning}
Large-scale recommendation systems face a similar scalability issue when storing embeddings for millions of categorical IDs. Deep Hash Embedding (DHE)~\cite{kang2020dhe} replaces explicit embedding tables with multiple universal hash functions followed by a compact MLP, computing embeddings on-the-fly and greatly reducing storage while mitigating hash collisions. The deformation features in dynamic 3DGS share a similar flavor: they live in a large, high-dimensional and potentially sparse domain, and naive table-based parameterizations quickly become memory-inefficient as spatial and temporal resolution increase. In practice, the dimensionality and density variability of deformation features make fixed hash tables particularly brittle, as they cannot adapt their capacity to local spatio-temporal complexity without incurring substantial overhead. Inspired by this learnable hash-table-free design philosophy, our method adopts a similar hash--MLP strategy for dynamic scene representation, replacing learnable multi-resolution hash grids for deformation features with a compact neural function that generates features at query time. This learnable hash-table-free deformation field serves as a storage-efficient counterpart to our subsequent compression of the canonical Gaussian point cloud.

%% file: Sections/Preliminaries.tex
\subsection{Principles of 3D Gaussian Splatting}
3DGS~\cite{3dgs} explicitly represents a scene using a collection of 3D Gaussian kernels, offering superior computational efficiency and interpretability compared to implicit models like NeRF~\cite{nerf}. Each Gaussian primitive is parameterized by a center position $\bm{\mu} \in \mathbb{R}^3$, a spatial covariance matrix $\Sigma \in \mathbb{R}^{3 \times 3}$, an opacity $o \in \mathbb{R}$, and a color vector $\mathbf{c}$ encoded via SH coefficients.
In practice, the covariance matrix $\Sigma$ is decomposed into a scaling matrix $S$ and a rotation matrix $R$ to ensure positive semi-definiteness, expressed as $\Sigma = R S S^\top R^\top$.
For rendering, each 3D Gaussian is projected onto the 2D image plane, and we denote by $\bm{\mu}_i^p$ and $\Sigma_i^p$ the center (mean) and covariance of its 2D footprint for the $i$-th primitive. The per-pixel opacity contribution of this primitive at pixel $p$ is modeled as
\begin{equation}
\alpha_i(p) = o_i \exp \left(
    -\frac{1}{2} (p - \bm{\mu}_i^p)^\top (\Sigma_i^p)^{-1} (p - \bm{\mu}_i^p)
\right).
\end{equation}
The final pixel color $C(p)$ is then computed via alpha-blending of the sorted primitives:
\begin{equation}
C(p) = \sum_{i=1}^{N} c_i \, \alpha_i(p) \prod_{j=1}^{i-1} \bigl(1 - \alpha_j(p)\bigr).
\end{equation}
where $\alpha_i$ represents the opacity of the $i$-th Gaussian after projection.

\subsection{Dynamic Scene Representation and Storage Challenges}
\label{sec:4dgs_problem}
To extend 3DGS to dynamic scenes, 4DGS frameworks typically define primitives in a canonical space that undergo time-varying deformation to match observed primitives in world space. Specifically, a canonical Gaussian $G$ is transformed into a deformed primitive $G'$ at time $t$ via a learned deformation $\Delta G$:
\begin{equation}
G' = G + \Delta G, \quad \text{where} \quad \Delta G = \Phi(f, t).
\end{equation}
Here, $\Phi$ is a deformation network that predicts offsets based on the primitive's spatio-temporal features $f$ and the current time $t$.
A critical limitation of recent approaches is their reliance on massive multi-resolution hash tables or voxel grids to store these features $f$. This dependence results in a dramatic increase in model size, often ranging from hundreds of megabytes to several gigabytes per scene. Such significant storage overhead imposes severe constraints on scalability, streaming, and deployment on memory-constrained edge devices.

%% file: Sections/Method.tex
\begin{figure}
    \centering
    \includegraphics[width=1.0\linewidth]{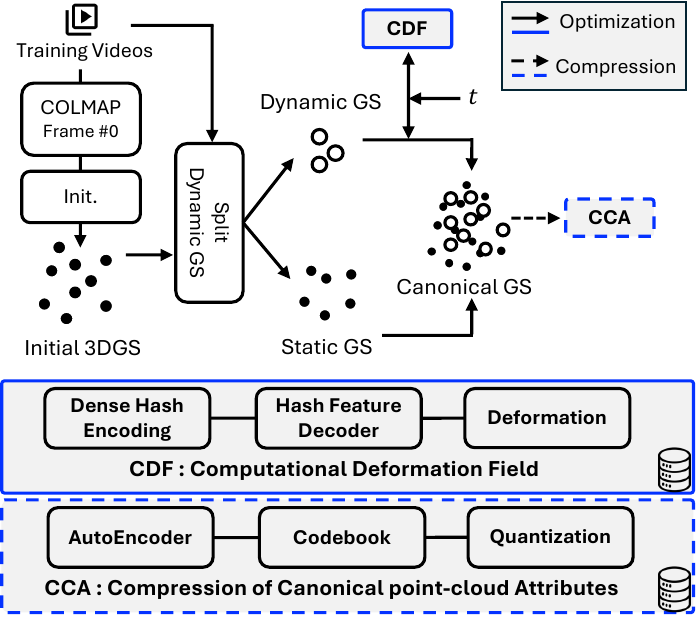}
    \caption{
          Overview of the proposed CC-4DGS framework.
          The pipeline begins with COLMAP pose estimation, followed by a coarse initialization (Init.) to generate the base Canonical Gaussians.
          These primitives are then separated into static and dynamic subsets (Dynamic GS Split).
          To minimize storage overhead, the CDF replaces learnable hash tables with a deterministic dense hash encoding and a Hash Feature Decoder, computing deformations on-the-fly during the main optimization.
          Finally, the CCA module compresses the optimized attributes via autoencoding and quantization (dashed lines). Disk icons indicate the final compressed components persisted in storage.
        }
\label{fig:framework}
\end{figure}
\subsection{Overview}
Fig.~\ref{fig:framework} shows the overall CC-4DGS framework. Starting from COLMAP~\cite{colmap} initialization, the scene is represented as canonical Gaussians, which are separated into static and dynamic subsets. Dynamic Gaussians are updated through the CDF, which predicts time-dependent deformation fields that warp each Gaussian from its canonical position to the target frame $t$. The deformation is computed on-the-fly using dense hash encoding, hash feature decoders, and a lightweight deformation network—avoiding the large learnable multi-resolution hash tables commonly used in prior methods and thereby enabling a far more storage-efficient deformation pathway (Sections IV-B and IV-C). The canonical Gaussian attributes are further compressed through the CCA module, which applies autoencoding, quantization, and residual codebooks to reduce both appearance and geometry storage (Section IV-D). Together, these components form a compact and scalable representation for dynamic 4D scenes.

\begin{figure*}
    \centering
    \includegraphics[width=1.0\linewidth]{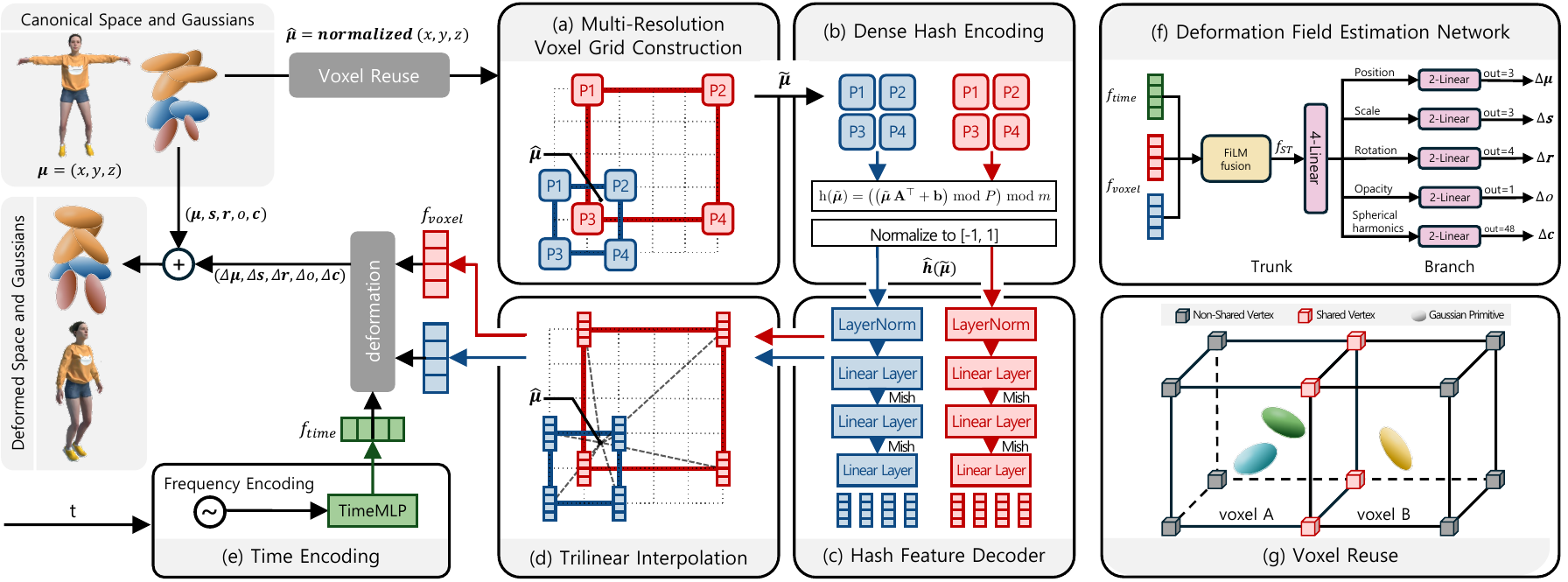}
    \caption{
        Overview of the proposed CDF.
        The normalized center $\hat{\bm{\mu}}$ is used to generate multi-resolution voxel vertices $\tilde{\bm{\mu}}$ (a), which are hashed and encoded into vectors $h(\tilde{\bm{\mu}})$ (b).
        These vectors are passed through resolution-specific Hash Feature Decoders (which map the DHE multi-hash signatures to deformation features rather than resolving collisions) to generate vertex features (c), which are interpolated to obtain the spatial feature $f_{\text{voxel}}$ (d).
        A separate temporal feature $f_{\text{time}}$ (e) is combined with $f_{\text{voxel}}$ and fed into the deformation network (f) to produce time-varying Gaussians. To reduce redundancy, the Voxel Reuse module (g) identifies shared vertices (red) between adjacent voxels to avoid re-computation.
        }
    \label{fig:flow}
\end{figure*}

\subsection{Computational Deformation Field}
\label{sec:cdf}
\subsubsection{\textbf{Learning Vertex Features without Hash Tables}}
\label{subsec:LearningVertexFeatuers}

The core challenge in designing a computational deformation field (synthesized per query by a neural decoder rather than retrieved from learnable tables) is to efficiently synthesize high-fidelity spatial features for every Gaussian without relying on large multi-resolution hash tables. To address this, we introduce the Hash Feature Decoder framework, which forms the central component of our CDF and enables on-the-fly computation of deformation fields. As illustrated in Fig.~\ref{fig:flow}, the pipeline proceeds through four integral stages: grid construction, hash encoding, neural decoding, and feature interpolation.

The process commences with the Multi-Resolution Voxel Grid Construction (Fig.~\ref{fig:flow}(a)). Given a primitive's center coordinate $\bm{\mu}\in\mathbb{R}^3$ in canonical space, we first normalize it to the unit cube $\hat{\bm{\mu}}\in[0,1]^3$. This continuous coordinate is discretized into an integer voxel index $\mathbf{v} = \lfloor R \cdot \hat{\bm{\mu}} \rfloor$ at a specific grid resolution $R$. To capture the local geometry surrounding the primitive, we generate the eight corner vertices $\tilde{\bm{\mu}}$ of the enclosing voxel by applying binary offsets to $\mathbf{v}$:
\begin{equation}
    \tilde{\bm{\mu}}_{(dx, dy, dz)} = \mathbf{v} + (dx, dy, dz), \quad \forall (dx, dy, dz) \in \{0, 1\}^3.
    \label{eq:vertex}
\end{equation}
In Fig.~\ref{fig:flow}(a), the vertices labeled P1 through P4 (colored in red and blue) visualize this multi-resolution spatial discretization.

Subsequently, we employ a Dense Hash Encoding strategy (Fig.~\ref{fig:flow}(b)) to map these spatial coordinates into a high-dimensional latent space. Instead of relying on a single lookup, each vertex coordinate $\tilde{\bm{\mu}}$ is projected through $k$ parallel universal hash functions. Parameterized by a matrix $\mathbf{A} \in \mathbb{Z}^{k \times 3}$ and a vector $\mathbf{b} \in \mathbb{Z}^{k}$, the hashing operation is formulated as:
\begin{equation}
\label{eq:each_hash_function}
h(\tilde{\bm{\mu}}) = \Bigl( \bigl(\tilde{\bm{\mu}} \mathbf{A}^\top +\mathbf{b} \bigr) \mod P \Bigr) \mod m,
\end{equation}
where $P$ is a large prime and $m$ denotes the modulo basis. The resulting integer vector is then normalized to $[-1, 1]^k$, yielding a dense, real-valued signature $\hat{h}(\tilde{\bm{\mu}})$.

This signature serves as the input to the Hash Feature Decoder as depicted in Fig.~\ref{fig:flow}(c). Departing from conventional methods that retrieve static features from explicit tables, our approach leverages a lightweight decoding network to synthesize vertex-specific features on-the-fly. Specifically, the decoder comprises a sequence of Linear Layers interleaved with Mish activation functions and LayerNorm, enabling a compact yet non-linear mapping of the hashed latent codes.

In the final stage, Trilinear Interpolation ensures spatial continuity (Fig.~\ref{fig:flow}(d)). The decoded features are mapped back to their corresponding grid vertices (visualized as colored feature vectors) and aggregated via weighted interpolation to derive the final spatial feature $f_{voxel}$ at the primitive center $\bm{\mu}$.

Our architecture differs from conventional grid-based hashing. By replacing large-scale hash tables with a dynamic Hash Feature Decoder, we effectively eliminate the memory bottleneck of explicit storage. Simultaneously, the DHE multi-hash construction reduces exact-signature collisions and the decoder maps the resulting high-dimensional signatures to continuous deformation features, thereby providing a favorable balance between storage efficiency, computational fidelity, and representational capacity.

\subsubsection{\textbf{Collision-Resilient Multi-Resolution Design}}
\label{subsec:collision-resilient}
Although multi-resolution grid representations have demonstrated substantial expressive capacity, they often incur prohibitive memory footprints due to their reliance on deep hierarchies (e.g., typically 16 levels in prior works~\cite{ingp, grid4d, Swift4d}). To address this limitation, we use the DHE multi-hash construction together with resolution-specific neural decoders to employ a streamlined voxel grid hierarchy comprising only 4 to 5 resolution levels.
By allocating a dedicated lightweight decoding network to each resolution, our framework effectively decouples the learning of global geometry from fine-grained details. This strategy ensures that the model maintains representational fidelity comparable to deeper hierarchies, while significantly optimizing computational and storage efficiency.

The collision-resilient behavior of this design is primarily supported by the DHE multi-hash construction. Exact ambiguity would require two distinct coordinates to share the same signature across all hash dimensions; in our actual N3DV query domain, the full ($k{=}128$) signatures produced zero exact duplicates across all queried voxel-corner coordinate sets. The Hash Feature Decoder then learns a joint mapping from these high-dimensional signatures to continuous deformation features, rather than explicitly resolving arbitrary hash collisions.
\subsubsection{\textbf{Time Encoding and Deformation Field Estimation}}
\label{subsec:time}

While the Hash Feature Decoder robustly encodes static spatial geometry, high-fidelity modeling of dynamic scenes necessitates the rigorous integration of temporal variations. To this end, we employ a frequency-based positional encoding followed by a lightweight temporal encoder to represent the input timestamp $t$ with high precision, as illustrated in Fig.~\ref{fig:flow}(e).
First, the scalar timestamp is expanded into a high-dimensional vector $\gamma(t)$ using sinusoidal functions:
\begin{equation}
\gamma(t) = \Bigl[ t, \bigl\{ \sin(2\pi f_l t), \cos(2\pi f_l t) \bigr\}_{l=1}^L \Bigr],
\label{eq:time_encoding}
\end{equation}
where $f_l=2^{l-1}$ represents the frequency of the $l$-th band, and $L$ denotes the total number of frequency bands. This encoding is subsequently processed by a shallow neural module (TimeMLP) to yield the high-level temporal feature $f_{\text{time}}$.

To effectively synthesize spatiotemporal context, we integrate this temporal feature $f_{time}$ with the spatial feature $f_{voxel}$ retrieved from the grid. Critically, rather than resorting to naive concatenation, we adopt a Feature-wise Linear Modulation (FiLM)~\cite{film} mechanism. This design ensures that temporal dynamics condition the underlying spatial structure explicitly.
We first project the voxel feature into a latent space via a linear transformation to obtain the spatial hidden state $h_{spatial}$:
\begin{equation}
h_{spatial} = W_s f_{voxel} + b_s.
\end{equation}
In parallel, a compact FiLM generator network takes $f_{time}$ as input and regresses the channel-wise scale $\gamma(f_{time})$ and shift $\beta(f_{time})$ vectors. The unified spatiotemporal feature is then computed via affine modulation:
\begin{equation}
f_{ST} = \gamma(f_{time}) \odot h_{spatial} + \beta(f_{time}),
\end{equation}
where $\odot$ denotes element-wise multiplication. By initializing $\gamma$ near unity and $\beta$ near zero, the network begins training as a quasi-static model, progressively learning to selectively amplify or suppress spatial channels to capture time-varying phenomena.

Subsequently, the fused feature $f_{ST}$ drives the Deformation Field Estimation Network (Fig.~\ref{fig:flow}(f)) to predict the time-dependent evolution of each Gaussian primitive. As depicted in the figure, the architecture follows a Trunk-Branch design: a shared trunk of four linear layers extracts global deformation context, which is then distributed to task-specific branches (prediction heads). Each branch consists of two linear layers to independently regress offsets for specific attributes (position $\Delta \bm{\mu}$, scale $\Delta \mathbf{s}$, rotation $\Delta \mathbf{r}$, opacity $\Delta o$, and SH coefficients $\Delta\mathbf{c}$). All intermediate layers utilize Rectified Linear Units (ReLU) for non-linearity.
This framework facilitates a continuous dynamic mapping from the canonical Gaussian space to the observation space as defined in Eq.~(\ref{eq:deformation}):
\begin{equation}
\Delta \theta = f_{\theta}(f_{\text{ST}}) \quad \text{for } \theta \in \{\bm{\mu}, \mathbf{s}, \mathbf{r}, o, \mathbf{c} \}.
\label{eq:deformation}
\end{equation}
These predicted offsets are applied additively to the canonical attributes, realizing smooth dynamic transitions in position, orientation, and appearance during the rasterization process.

\subsection{Overcoming the Computational Overhead of CDF}
\label{sec:fps_strategy}

\subsubsection{\textbf{Vertex Reuse for Redundant-Compute Reduction}}
\label{subsec:vertex_reuse}

While the proposed CDF eliminates the massive storage overhead of the large learnable multi-resolution hash tables used by prior methods by synthesizing features on-the-fly, this procedural generation introduces a computational cost.
Specifically, as detailed in the grid construction phase, multiple Gaussian primitives residing within the same or adjacent voxels inevitably reference identical vertex coordinates $\tilde{\bm{\mu}}$. Naive execution would thus trigger redundant hash encoding and neural decoding operations for these shared vertices.
To mitigate this computational redundancy, we implement a Vertex Reuse mechanism as illustrated in Fig.~\ref{fig:flow}(g).
Prior to feature inference, we aggregate all query vertex coordinates $\tilde{\bm{\mu}}$ derived from the normalized Gaussian centers $\hat{\bm{\mu}}$ and filter them to construct a unique vertex set.
The decoding network is then evaluated exclusively on this unique set.
Subsequently, each Gaussian retrieves its required vertex features via index-based lookups from this precomputed buffer, followed by trilinear interpolation.
By maintaining a lightweight mapping between primitives and unique vertices in memory, we ensure efficient retrieval without re-computation.
This strategy is particularly advantageous in high-resolution voxel grids, where the ratio of Gaussians to unique vertices is high.
By effectively amortizing the inference cost across spatially adjacent primitives, the Vertex Reuse module significantly reduces the computational load and GPU memory footprint, thereby enhancing the real-time rendering frame rate.

\subsubsection{\textbf{Spatial-Temporal Feature Caching}}
\label{subsec:feature_caching}

Although the Vertex Reuse strategy minimizes redundant computations within a single frame, the naive re-evaluation of the spatial decoder for every frame remains a bottleneck compared to the $O(1)$ lookup complexity of conventional hash tables.
To bridge this performance gap, we introduce a Spatiotemporal Feature Caching scheme that exploits the temporal invariance of the canonical space.
Our framework inherently decouples the feature generation process into a spatial component $f_{\text{voxel}}$ and a temporal component $f_{\text{time}}$.
Since the canonical Gaussian primitives are static, their associated spatial features $f_{\text{voxel}}$ remain constant throughout the sequence.
Conversely, the temporal feature $f_{\text{time}}$ is dynamic and strictly dependent on the timestamp $t$.
Leveraging this property, we compute the spatial features $f_{\text{voxel}}$ only during the initialization phase (or the first frame) and cache the results in a persistent buffer, denoted as $f_{\text{voxel\_cached}}$.
For all subsequent frames, the system bypasses the spatial decoder entirely, retrieving $f_{\text{voxel\_cached}}$ from memory and fusing it with the newly computed $f_{\text{time}}$.
This approach effectively eliminates the overhead of repetitive spatial decoding while preserving the flexibility to model complex temporal dynamics, resulting in a substantial improvement in rendering speed without compromising visual fidelity.

\subsection{Compression of Canonical Point-Cloud Attributes}
\label{subsec:pointcloud_compression}

\begin{figure}[t]
    \centering
    \includegraphics[width=1.0\linewidth]{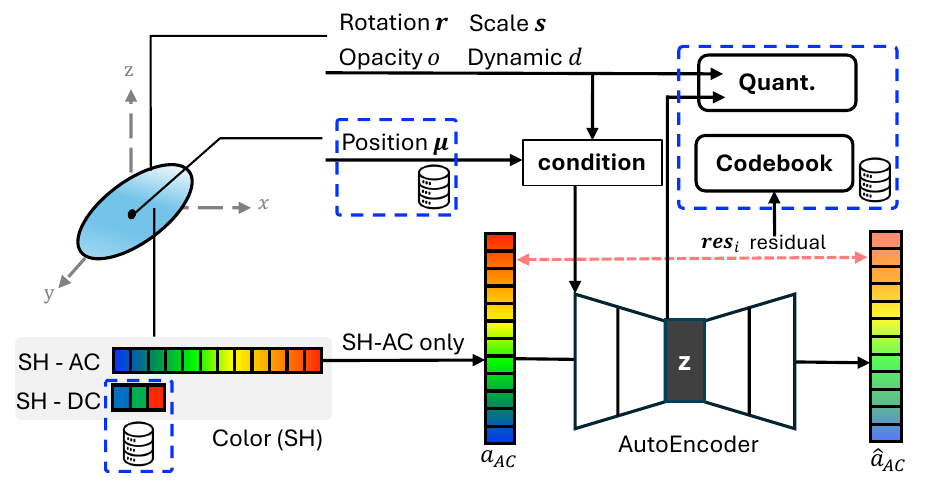}
    \caption{
          Pipeline of the proposed CCA.
          To optimize storage efficiency while maintaining visual fidelity, Gaussian attributes are categorized and processed via three distinct pathways.
          (Left) Crucial attributes, including Position ($\bm{\mu}$) and SH-DC components, are retained in full precision to prevent geometric collapse.
          (Top) Scalar attributes ($\mathbf{r}, \mathbf{s}, o, d$) undergo Selective Quantization, where dynamic logits $d$ are specifically pre-processed via $\mu$-law companding.
          (Bottom) High-dimensional SH-AC features are compressed by a Conditional Autoencoder that leverages geometric context (condition block) to maximize encoding efficiency.
          Residual errors ($\mathbf{res}_i$) are further captured by a Vector Quantization Codebook.
          Blue dashed boxes and disk icons indicate the final compressed components stored in the archive.
        }
\label{fig:pcc}
\end{figure}

Although the proposed CDF substantially improves the storage efficiency of the deformation field, the canonical point cloud—consisting of millions of primitives—still dominates the overall memory footprint. To address this issue, we design a hybrid compression framework that accounts for the sensitivity and dimensional characteristics of each attribute group. CCA combines well-established compression primitives---autoencoding, scalar quantization, and codebook-based residual coding---into an attribute-aware pipeline tailored to the canonical Gaussian set of our dynamic 4DGS framework.
Fig.~\ref{fig:pcc} illustrates the overall processing pipeline of the proposed CCA. The input Gaussian attributes are categorized into three groups based on their characteristics and processed through separate pathways.
First, spatial positions ($\bm{\mu}$) and the DC component of the spherical harmonics (SH-DC), which are highly sensitive to information loss and critical for visual fidelity, are preserved without compression and stored directly, as indicated by the blue dashed box.
Second, the remaining scalar attributes—rotation $\mathbf{r}$, scale $\mathbf{s}$, opacity $o$, and the dynamic indicator $d$—are routed to the quantization module along the upper path of the figure.
Third, the AC component of the spherical harmonics (SH-AC), which account for the majority of the storage cost, are processed through the lower path comprising an autoencoder and a codebook-based residual representation. A central condition block aggregates geometric attributes and injects them into the autoencoder, enabling the model to exploit correlations between geometry and appearance during compression.

The detailed mechanisms of each processing pathway are described next.

\subsubsection{\textbf{Compression of SH-AC Appearance Components}}
As illustrated along the lower pathway of Fig.~\ref{fig:pcc}, the SH-AC components, which account for approximately 70\% of the total data size, are compressed using a conditional autoencoder. Instead of using only the SH-AC vector $\mathbf{a}_{\mathrm{AC}}$ as input, we construct a joint input vector $\mathbf{u} = [\,\mathbf{a}_{\mathrm{AC}},\, \mathbf{y}\,]$ by concatenating the SH-AC features with the conditioning attributes $\mathbf{y} = (\bm{\mu}, \mathbf{s}, o, \mathbf{r}, d)$ comprising the per-Gaussian position, scale, opacity, rotation, and dynamic indicator, as shown in the central portion of the figure. The encoder maps $\mathbf{u}$ into a low-dimensional latent vector $\mathbf{z}$, which is subsequently quantized and stored as a compact bitstream. This conditioning mechanism enables the model to capture correlations between geometry and appearance, thereby improving compression efficiency.
To further correct reconstruction errors, a codebook module, shown on the right side of the figure, is employed. The residual vector $\mathbf{res}_i$ between the original and reconstructed SH-AC features is computed and then encoded using K-means–based vector quantization with a $256$-entry residual codebook, storing the codebook and one assignment index per Gaussian. This two-stage design allows the system to recover fine-grained texture details with minimal additional storage overhead.

\subsubsection{\textbf{Quantization of Geometry and Scalar Attributes}}
For the attributes routed through the upper path of Fig.~\ref{fig:pcc}, a selective quantization strategy is applied based on their sensitivity. As noted earlier, the spatial positions $\bm{\mu}$ are stored with full 32-bit precision in order to prevent geometric distortions. In contrast, rotation, scale, and opacity are less sensitive to quantization errors and are therefore compressed using 12-bit linear quantization.
The dynamic indicator $d$ exhibits a long-tailed distribution, and simple linear quantization is insufficient to preserve its fine-grained variations. To address this issue, we apply $\mu$-law companding, defined in \eqref{eq:mulaw}:
\begin{equation}
\tilde{d} = \operatorname{sign}(d)\, \frac{\ln(1 + \mu |d|)}{\ln(1 + \mu)}.
\label{eq:mulaw}
\end{equation}
This transformation allocates higher effective resolution to the region where important information is concentrated.
Finally, the outputs from all processing paths are combined with lightweight metadata and stored after lossless compression using Zstandard~\cite{collet2016zstd}, as indicated by the blue dashed box in the figure.

\subsection{Loss Functions and Regularization}
During training, at each iteration, the predicted deformations are applied to each Gaussian’s attributes, including position, scale, rotation, opacity, and SH coefficients. The rendered output is supervised using a standard pixel-wise $L_1$ reconstruction loss and SSIM loss.
However, despite accurate deformation prediction by our model, we still observe abrupt changes in the deformation fields of explicit voxel-based representations for a small subset of scenes. To address this, inspired by Grid4D~\cite{grid4d}, we introduce a smoothness regularization loss applied directly to the embedded feature vectors prior to the deformation network inference, enhancing efficiency. 
In practice, we observe such artifacts only in a few challenging scenes, and thus enable the smoothness regularization solely for those scenes that exhibit noticeable temporal flickering.
Specifically, for a Gaussian primitive located at canonical coordinates $\bm{\mu}=(x,y,z)$ and timestamp $t$, we randomly sample small spatial and temporal perturbations  
$\boldsymbol{\epsilon}=(\epsilon_x,\epsilon_y,\epsilon_z)$ and $\epsilon_t$, drawn uniformly from $[-\delta,\delta]$ with $\delta=0.005$.  
We denote the concatenated deformation outputs derived from embedded features as $\mathbf{d}$ and $\mathbf{d}'$ for original and perturbed inputs, respectively. Our smoothness regularization loss then penalizes the difference between these outputs:
\begin{equation}
\mathcal{L}_{\text{smooth}} = \|\mathbf{d}-\mathbf{d}'\|_2^2
\label{eq:Lr}
\end{equation}
The final loss function combines reconstruction accuracy and smoothness regularization:
\begin{equation}
\mathcal{L}_{\text{total}} = \lambda_c\mathcal{L}_1 + (1-\lambda_c)(1-\mathcal{L}_{SSIM}) + \lambda_{r}\mathcal{L}_{\text{smooth}},
\end{equation}
where $\lambda_{r}$ balances between the fidelity of reconstruction and the smoothness of the deformation fields. 
The weight $\lambda_{r}$ is linearly increased as training progresses.
By enforcing similarity among deformation parameters within local spatio-temporal neighborhoods, this regularization effectively suppresses flickering artifacts and stabilizes the rendering of dynamic regions.

%% file: Sections/Performance_Evaluation.tex
\subsection{Experimental Environment} 
\label{sec:exp_settings}
\subsubsection{Experimental Configuration} 
To evaluate the performance of CC-4DGS, we compare it against representative dynamic Gaussian splatting methods, including 4DGS~\cite{4dgs_wu}, 4D-GS~\cite{4dgs_yang}, E-D3DGS~\cite{e-d3dgs}, and the hash-table–based approaches Grid4D~\cite{grid4d} and Swift4D~\cite{Swift4d}. For a fair assessment of storage efficiency, point cloud compression using the CCA module is applied equally to Swift4D and our method, enabling comparison under identical compressed-geometry conditions.
All experiments follow a single-step initialization protocol based on a single canonical point cloud. COLMAP~\cite{colmap} is applied only to the first frame to obtain initial geometry, and subsequent deformations are learned solely from 2D supervision. This setting isolates the effect of removing large hash tables, ensuring an unbiased comparison. Methods such as SaRO-GS, SpaceTime-GS, Ex4DGS~\cite{sarogs,spacetimegs,ex4dgs} and streaming-oriented approaches including 4DGC, 4DGV, and 3DGStream~\cite{4dgc,4dgv,3dgstream} rely on continuous geometric priors and therefore operate under different assumptions; they are excluded from our main comparison.
All baseline methods are evaluated using their official implementations. Pretrained weights are used when available, and otherwise models are retrained following their official configurations.
These restrictions apply to the direct head-to-head comparisons in Table~\ref{tab:neu3d} and Table~\ref{tab:tech};  for additional context, the lower block of Table~\ref{tab:neu3d} compiles reference values from recent compression-oriented 4DGS work using each method's own reported numbers, including methods (e.g., the streaming-task 4DGC) that fall outside the protocol above.
\subsubsection{Datasets and Evaluation Metrics} 
The evaluation uses two real-world datasets containing dynamic scenes: Neural 3D Video (N3DV)~\cite{neu3d} and Technicolor Light Field~\cite{technicolor}. N3DV provides multiview sequences at $2704 \times 2028$ and 30 frames per second (FPS), captured by 18–21 cameras over roughly 10 seconds. Following prior work, all videos are downsampled to half resolution, and views with severe blur artifacts are excluded.
The Technicolor dataset contains dynamic scenes recorded with a $4 \times 4$ camera array, each providing synchronized $2048 \times 1044$ video. As in E-D3DGS~\cite{e-d3dgs}, the central view is used, and five representative scenes (\texttt{Birthday}, \texttt{Fabien}, \texttt{Painter}, \texttt{Theater}, \texttt{Trains}) with 50 frames each are selected for training and evaluation.
Quantitative evaluation uses PSNR and SSIM, while rendering speed is measured in FPS on a single NVIDIA RTX~4090 GPU. Storage cost is reported separately for the deformation module and the point cloud to highlight their respective contributions to total model size.
\subsubsection{Implementation Details for Evaluation} 
All experiments are implemented in PyTorch~\cite{pytorch} and conducted on a single NVIDIA RTX 4090 GPU. Hyperparameters follow those of 4DGS and Swift4D, adjusted to fit our model architecture.
Unlike conventional hash-grid approaches that store vertex embeddings in large XOR-based hash tables, CC-4DGS stores no such learnable hash tables to reduce storage. We employ modulation-based universal hash functions with dense hash encoding, and use lightweight neural decoders at each grid level to map hash-encoded integers to continuous features. Each grid level uses a sufficiently large bucket size $m$ and a prime $P$ to limit hash collisions, as preliminary tests showed that smaller $m$ values noticeably degrade PSNR. Due to GPU memory constraints, the number of multi-resolution grids is restricted to four or five.
We also incorporate the dynamic-score strategy from Swift4D, applying deformation only to Gaussians whose learned dynamic score $d$ exceeds a threshold $\zeta$, which avoids unnecessary computation in static regions and improves training and rendering efficiency.
CC-4DGS follows a three-stage schedule: $5$K iterations for coarse initialization, $3$K iterations for fitting the per-Gaussian dynamic score $d$, and $20$K iterations for the main optimization stage where the deformation field is jointly optimized with the Gaussian set ($28$K total per N3DV scene). Swift4D shares the same coarse and dynamic-score stages but uses its official $13$K main-stage configuration, and we account for this schedule difference in the optimization-cost analysis (Sec.~VI-E).

\subsection{Quantitative Results} 
Quantitative results on the N3DV dataset are presented in Table~\ref{tab:neu3d}. In terms of reconstruction quality, the proposed method achieves slightly higher performance than Swift4D~\cite{Swift4d}, reaching PSNR values in the low 32 dB range and an SSIM of approximately 0.94, placing it among the top-performing 4D Gaussian–based methods.
Although Swift4D attains higher FPS due to its 4D grid–based hash table and selective processing of dynamic Gaussians, the CDF pathway in our framework maintains real-time rendering performance through the combination of feature caching and dynamic separation. The specific impact of feature caching is analyzed in Sec.~\ref{subsec:effect_featurecaching}.
In terms of storage, using the CDF component alone reduces model size to a few tens of megabytes, significantly smaller than Swift4D. When combined with CCA, the total storage further decreases to approximately 30 MB, with only 0.01–0.02 dB of PSNR loss. This result is also notable when compared to the lightweight baseline 4DGS~\cite{4dgs_wu}, which achieves around 31.6 dB PSNR while requiring a larger model size. A detailed analysis of the storage–quality trade-off is provided in Sec.~\ref{sec:neu_rate_distortion}.

For the Technicolor Light Field dataset, Table~\ref{tab:tech} provides a summary of the quantitative comparison. Since E-D3DGS~\cite{e-d3dgs} also initializes a canonical point cloud using COLMAP at $t{=}0$, it serves as the primary baseline. The CDF configuration of our method achieves approximately 0.5 dB higher PSNR (from 32.83 to 33.29 dB), improved SSIM (from 0.900 to 0.912), and a substantial increase in rendering speed (from 46.2 to 64.7 FPS) compared to E-D3DGS.
When combined with CCA, the full system further reduces the model size to around 32 MB while maintaining virtually unchanged quality, achieving PSNR 33.3 dB, SSIM 0.912, and roughly 64 FPS.
These results indicate that, even with the additional decoder computation, CC-4DGS effectively leverages point cloud compression to deliver higher PSNR than the available E-D3DGS baseline while substantially lowering storage cost on the Technicolor dataset.

Across both the N3DV and Technicolor datasets, the proposed CC-4DGS consistently demonstrates a superior trade-off among storage efficiency, reconstruction quality, and rendering speed under the shared setting of using the same initial COLMAP prior. In particular, the full configuration with CDF+CCA provides comparable reconstruction quality to representative methods such as 4DGS, Swift4D, and E-D3DGS, while maintaining a model size of only a few tens of megabytes, establishing it as a highly storage-efficient representation.

\begin{table}[h]
\centering
\caption{Quantitative comparison on the N3DV dataset. The upper block reports methods evaluated under our single-step protocol (Sec.~\ref{sec:exp_settings}) at $1352\times1014$; within this block the best~(\textbf{bold}) and second-best~(\underline{underlined}) values in each column are highlighted. The lower block lists recent 4DGS compression/streaming methods using values reported in their respective papers under heterogeneous settings (resolution, GPU, task). These values serve as contextual references rather than a direct head-to-head comparison and are excluded from the best/second-best highlighting.}
\label{tab:neu3d}
\setlength{\tabcolsep}{3pt}
\resizebox{\columnwidth}{!}{%
\begin{tabular}{lcccc}
\toprule
\textbf{Method} &
\textbf{PSNR(dB)$\uparrow$} & \textbf{SSIM$\uparrow$} & \textbf{FPS$\uparrow$} & \textbf{Storage(MB)$\downarrow$} \\
\midrule
\multicolumn{5}{l}{\textit{Evaluated in our environment (RTX~4090, $1352\times1014$)}}\\
4D-GS~\cite{4dgs_yang}        & 31.39 & \textbf{0.944} & 105.3 & 2193 \\
4DGS~\cite{4dgs_wu}           & 31.58 & 0.937 & 126.8 & \underline{41} \\
E-D3DGS~\cite{e-d3dgs}        & 30.85 & 0.941 & 78.2  & 56 \\
Grid4D~\cite{grid4d}          & 31.20 & 0.937 & 111.9 & 280 \\
Swift4D~\cite{Swift4d}        & \underline{32.06} & \textbf{0.944} & \underline{136.2} & 146 \\
Swift4D(+CCA)                 & 32.03 & \textbf{0.944} & \textbf{138.2} & 94 \\
\textbf{Ours(CDF)}            & \textbf{32.07} & \underline{0.943} & 111.9 & 86 \\
\textbf{Ours(CDF+CCA)}        & \underline{32.06} & \underline{0.943} & 113.4 & \textbf{30} \\
\midrule
\multicolumn{5}{l}{\textit{Reported values from recent methods (heterogeneous settings; not re-run)}}\\
Light4GS~\cite{light4gs}$^{\S}$ & $31.69$ & \hphantom{$^{\P}$}---$^{\P}$ & \hphantom{$^{\dagger}$}$37^{\dagger}$ & $5$ \\
GIFStream~\cite{gifstream}     & $31.75$          & $0.938$ & $95$  & $10$ \\
MEGA~\cite{mega}               & $31.49$          & \hphantom{$^{\P}$}$0.942^{\P}$ & \hphantom{$^{\dagger}$}$77.4^{\dagger}$ & $25$ \\
4DGC~\cite{4dgc}$^{\star}$     & $31.58$          & $0.943$ & \hphantom{$^{\dagger}$}$168^{\dagger}$  & $150$ \\
\bottomrule
\end{tabular}%
}
\\[0.3em]
\begin{minipage}{\linewidth}
\scriptsize
Recent-method rows are reproduced from each paper's reported numbers (no re-runs). Unless noted, recent methods report at $1352\times1014$, matching our setting. $^{\S}$ Light4GS is evaluated at $1024\times768$. $^{\P}$ SSIM values use the same single-scale definition as ours; MEGA's reported DSSIM$_1$ is converted to SSIM, whereas Light4GS reports only an MS-SSIM--based D-SSIM and is therefore omitted (---). $^{\dagger}$ FPS measured on a different GPU than our RTX~4090; not directly comparable (unmarked FPS use the same RTX~4090 as ours). $^{\star}$ Streaming-task method; storage is computed from the reported per-frame bitrate ($\sim$$0.5$ MB/frame) over a $300$-frame N3DV sequence.
\end{minipage}
\end{table}
\begin{table}[h]
\centering
\caption{Quantitative comparison of rendering quality, speed, and storage on the Technicolor dataset ($2048\times1044$ resolution). 
The \textbf{best} results for each column are highlighted.}
\label{tab:tech}
\setlength{\tabcolsep}{3pt}
\begin{tabular}{lcccc}
\toprule
\textbf{Model} & 
\textbf{PSNR(dB)$\uparrow$} & 
\textbf{SSIM$\uparrow$}& 
\textbf{FPS$\uparrow$} & 
\makecell{\textbf{Storage(MB)$\downarrow$}}\\
\midrule
E-D3DGS~\cite{e-d3dgs} & 32.83 & 0.900 & 46.2 & 58 \\
\textbf{Ours(CDF)} & 33.29 & \textbf{0.912} & \textbf{64.7} & 93 \\
\textbf{Ours(CDF+CCA)}& \textbf{33.30}& \textbf{0.912} & 64.6 &  \textbf{32}\\
\bottomrule
\end{tabular}
\end{table}

 The lower block of Table~\ref{tab:neu3d} additionally situates CC-4DGS among recent 4DGS compression and streaming methods that report N3DV numbers under their own evaluation settings. Because resolution, GPU, and task formulation differ across these methods, this block is a context-level comparison rather than a direct head-to-head benchmark: it shows that CC-4DGS occupies a high-fidelity operating point, reporting higher PSNR ($32.06$~dB) than the listed recent baselines at the cost of larger total storage than bitrate-oriented methods such as GIFStream ($10$~MB) and Light4GS ($5$~MB). It is therefore best read as a storage--quality--speed operating-point comparison rather than a uniform-dominance claim.

\subsection{Qualitative Results}
\label{sec:qualitative}

\begin{figure*}[t]
\centering
\includegraphics[width=1.0\linewidth]{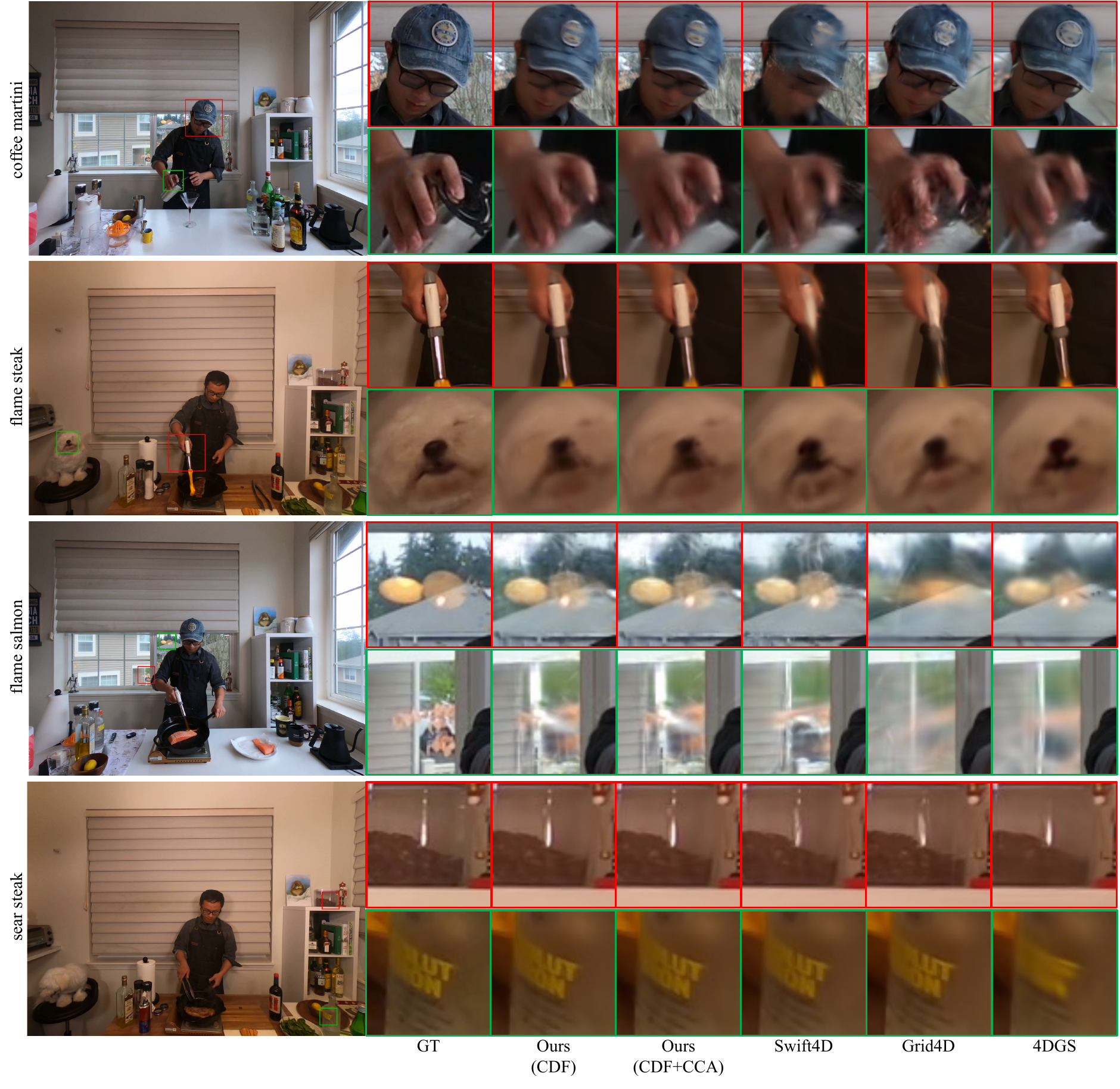}
\caption{
Qualitative comparison on the N3DV dataset. We compare rendering results for representative scenes (\texttt{coffee\_martini}, \texttt{flame\_steak}, \texttt{flame\_salmon}, \texttt{sear\_steak}) 
across the ground truth (GT), the proposed method (\textit{Ours(CDF)}, \textit{Ours(CDF+CCA)}), Swift4D~\cite{Swift4d}, Grid4D~\cite{grid4d}, and 4DGS~\cite{4dgs_wu}. 
For each scene, the left column shows the full-frame image, while the right column presents magnified regions containing high-frequency details and rapid motion, 
such as facial features, fingers, flame boundaries, and bottle labels, to highlight fine-grained differences among the methods. Despite the additional point-cloud compression, 
\textit{Ours(CDF+CCA)} maintains visual quality that is nearly indistinguishable from \textit{Ours(CDF)} to the human eye.}
\label{fig:n3dv_qual}
\end{figure*}

\begin{figure*}[t]
\centering
\includegraphics[width=1.0\linewidth]{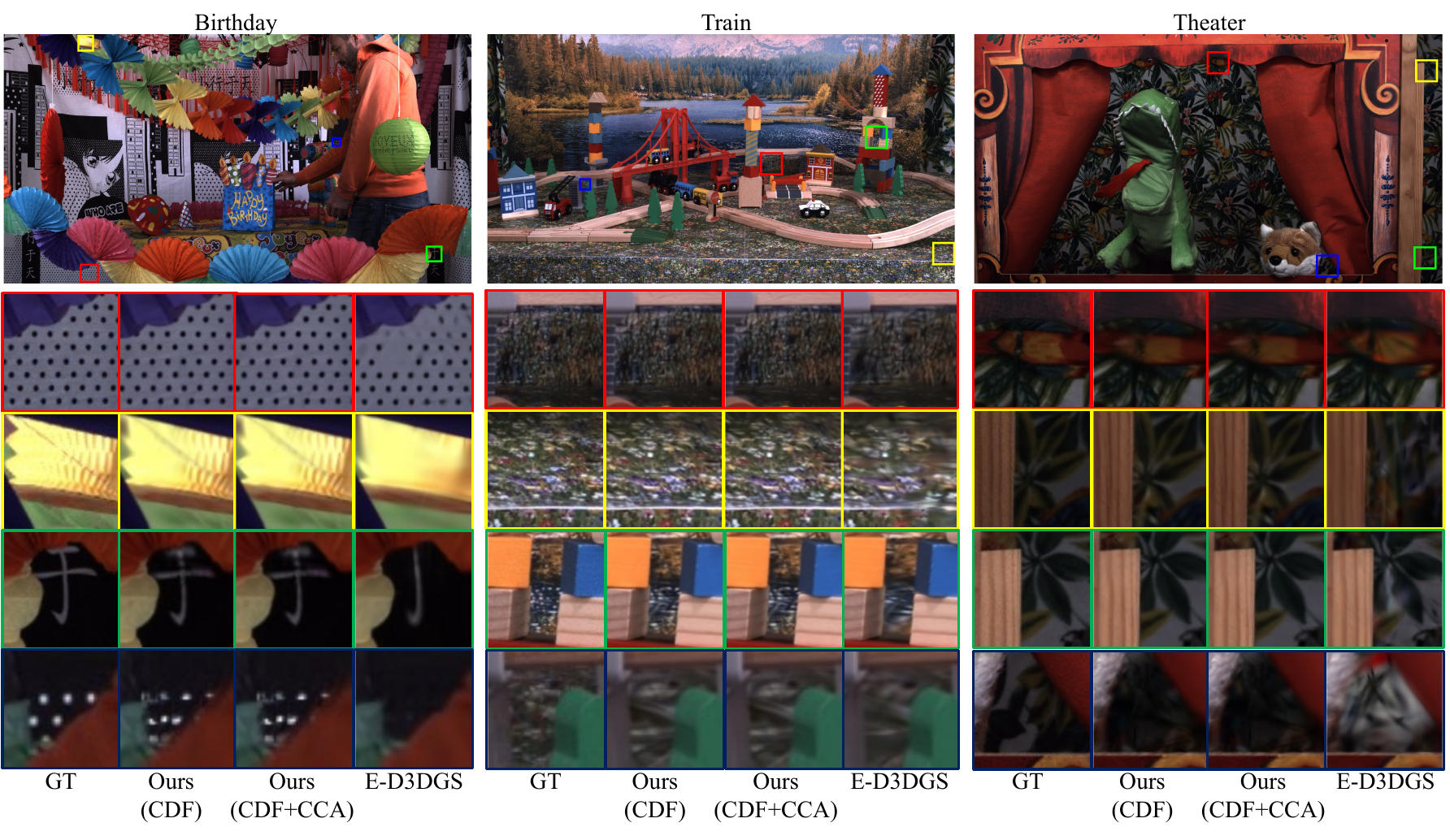}
\caption{Qualitative comparison on the Technicolor Light Field dataset. We compare the rendering results of the GT, the proposed method (\textit{Ours(CDF)}, \textit{Ours(CDF+CCA)}), and E-D3DGS across the \texttt{Birthday}, \texttt{Theater}, and \texttt{Train} scenes. To assess texture preservation and geometric stability, we highlight regions containing complex textures and thin structures—such as paper decorations, fabric folds, leaves, and building surfaces—as well as areas exhibiting large depth parallax, and present magnified views for detailed inspection.}
\label{fig:technicolor_qual}
\end{figure*}

We conduct qualitative comparisons on representative scenes from the N3DV and Technicolor Light Field datasets.
Figs.~\ref{fig:n3dv_qual} and \ref{fig:technicolor_qual} present side-by-side visual results for each scene, including the ground truth (GT), 4DGS~\cite{4dgs_wu}, Grid4D~\cite{grid4d}, Swift4D~\cite{Swift4d}, E-D3DGS~\cite{e-d3dgs}, and our method (\textit{Ours(CDF)} and \textit{Ours(CDF+CCA)}). Specifically, regions containing high-frequency textures or thin structures—such as hat logos, fingers, flames, paper decorations, and leaves—as well as areas involving fast motion are magnified to reveal fine-grained differences among the methods.

\noindent\textbf{Visual fidelity and detail preservation}
As shown in Fig.~\ref{fig:n3dv_qual}, existing dynamic 3DGS methods often exhibit over-smoothing in high-frequency regions, resulting in blurred edges and loss of fine structures. For example, in the \texttt{coffee\_martini} scene, both the hat logo and finger contours appear noticeably degraded in 4DGS and Grid4D, whereas CC-4DGS preserves sharp edges and textures close to those of the ground truth. A similar trend is observed in the \texttt{flame\_steak} scene, where our method reconstructs thin and semi-transparent flame structures more clearly, while Swift4D and Grid4D produce foggy artifacts around the flames.
These differences are consistently observed in \texttt{flame\_salmon} and \texttt{sear\_steak}. In \texttt{flame\_salmon}, prior methods fail to maintain the layered structure and silhouettes of distant background elements, whereas CC-4DGS preserves their separation more faithfully. In \texttt{sear\_steak}, existing approaches tend to blur specular highlights or distort label edges on reflective objects, while CC-4DGS more accurately reproduces highlight shapes and texture details, resulting in clearer detail in these examples.

\noindent\textbf{Complex backgrounds and high-resolution details.}
The Technicolor Light Field dataset contains higher-resolution imagery with numerous fine-scale structures across the scene, making even small reconstruction errors more noticeable. As shown in the \texttt{Birthday} and \texttt{Theater} scenes in Fig.~\ref{fig:technicolor_qual}, our method preserves thin structures and intricate patterns—such as paper decorations, fabric folds, and stage ornaments—with clear definition even at high resolution.
In contrast, E-D3DGS often exhibits over-smoothing in the same regions, where textures blend together or fine patterns disappear. Similar differences appear in the \texttt{Train} scene: our method maintains distinguishable details in high-frequency regions such as grass blades and brick surfaces, whereas E-D3DGS tends to blur or merge these patterns, resulting in a simplified appearance. These observations suggest that CC-4DGS more faithfully preserves fine textures and structural details in high-resolution environments like Technicolor.

\noindent\textbf{Effect of CCA on Visual Quality.}
As shown in Fig.~\ref{fig:n3dv_qual} and Fig.~\ref{fig:technicolor_qual}, \textit{Ours(CDF+CCA)} maintains visual quality nearly identical to that of \textit{Ours(CDF)} across most scenes. This is because the canonical geometry and appearance attributes are separated and re-encoded in a compression-friendly manner, allowing essential visual cues to be preserved even under lossy compression. Consequently, storage can be substantially reduced with negligible quality degradation, demonstrating that CC-4DGS achieves an effective trade-off between reconstruction quality and storage cost.

\subsection{Storage and Rate Distortion Analysis}
\label{sec:neu_rate_distortion}
This section analyzes the storage distribution and the resulting rate–distortion trade-off to demonstrate why the proposed CDF-based deformation representation is significantly more storage-efficient than approaches relying on large 4D hash tables. Table~\ref{tab:neu3dstorage} compares model size on the N3DV dataset by separating it into the \textit{Deform} module and the \textit{Point Cloud}. The \textit{Deform} component includes the parameters responsible for predicting time-varying properties such as position, scale, and opacity, corresponding to large multi-resolution hash grids in Swift4D~\cite{Swift4d} and to the CDF module in our method. The \textit{Point Cloud} contains canonical geometry and appearance attributes.
As shown in the table, most storage differences arise from the Deform module. Grid4D and Swift4D must store up to 16 levels of 3D/4D hash grids, resulting in Deform module sizes of 221.1 MB and 68.4 MB, respectively. In contrast, CC-4DGS eliminates the large learnable hash tables of prior methods and instead uses lightweight neural decoders to map dense hash encodings at each grid level, reducing the Deform module size to approximately 1.9 MB.
For the Point Cloud, applying the same CCA compression to all methods leads to similar storage usage in the tens of megabytes range. Consequently, differences in total storage stem primarily from the deformation design. Notably, \textit{Ours(CDF+CCA)} maintains more than a 60 MB reduction in total size compared to Swift4D+CCA and Grid4D.

To further examine differences within the \textit{Deform} module, Table~\ref{tab:neu3dstorage_breakdown} decomposes its storage into \textit{Deform-Grid} and \textit{Deform-Net}. For Grid4D and Swift4D, the majority of the storage is consumed by the multi-resolution 3D/4D hash grids in Deform-Grid, while the parameters used to predict deformations (Deform-Net) remain below 1 MB. In contrast, the proposed method eliminates the large learnable hash tables of prior methods, reducing Deform-Grid to approximately 0.8 MB and allocating the remaining ~1.1 MB to a compact neural network. This indicates that prior methods devote most of their bit budget to storing large lookup tables, whereas our design reallocates the same order of megabytes toward lightweight learned parameters.
\begin{table}[h]
\centering
\caption{Storage breakdown on the N3DV dataset. The total model size is decomposed into the \textit{Deform} module and the \textit{Point Cloud} components. The \textit{Deform} module includes parameters for predicting time-varying Gaussian attributes, while the \textit{Point Cloud} corresponds to canonical geometry and appearance storage. The \textbf{Best} results for each column are highlighted.}
\label{tab:neu3dstorage}
\setlength{\tabcolsep}{3pt}
\begin{tabular}{l
                S[table-format=3.1]
                S[table-format=3.1]
                S[table-format=3.1]}
\toprule
\textbf{Model} & 
\multicolumn{1}{c}{\makecell{\textbf{Size(MB)$\downarrow$}\\\textbf{(Deform)}}} &
\multicolumn{1}{c}{\makecell{\textbf{Size(MB)$\downarrow$}\\\textbf{(Point Cloud)}}} &
\multicolumn{1}{c}{\makecell{\textbf{Size(MB)$\downarrow$}\\\textbf{(Total)}}} \\
\midrule
Grid4D~\cite{grid4d}      & 221.1 &  58.8 & 279.9 \\
Swift4D~\cite{Swift4d}    &  68.4 &  77.4 & 145.8 \\
Swift4D(+CCA)              &  68.4 &  \bfseries 25.7 &  94.1 \\
\midrule
\textbf{Ours(CDF)}             & \bfseries 1.9 &  84.4 &  86.4 \\
\textbf{Ours(CDF+CCA)}        & \bfseries 1.9 &  27.7 &  \bfseries 29.6 \\
\bottomrule
\end{tabular}
\end{table}

\begin{figure}[h]
\centering
\includegraphics[width=1.0\linewidth]{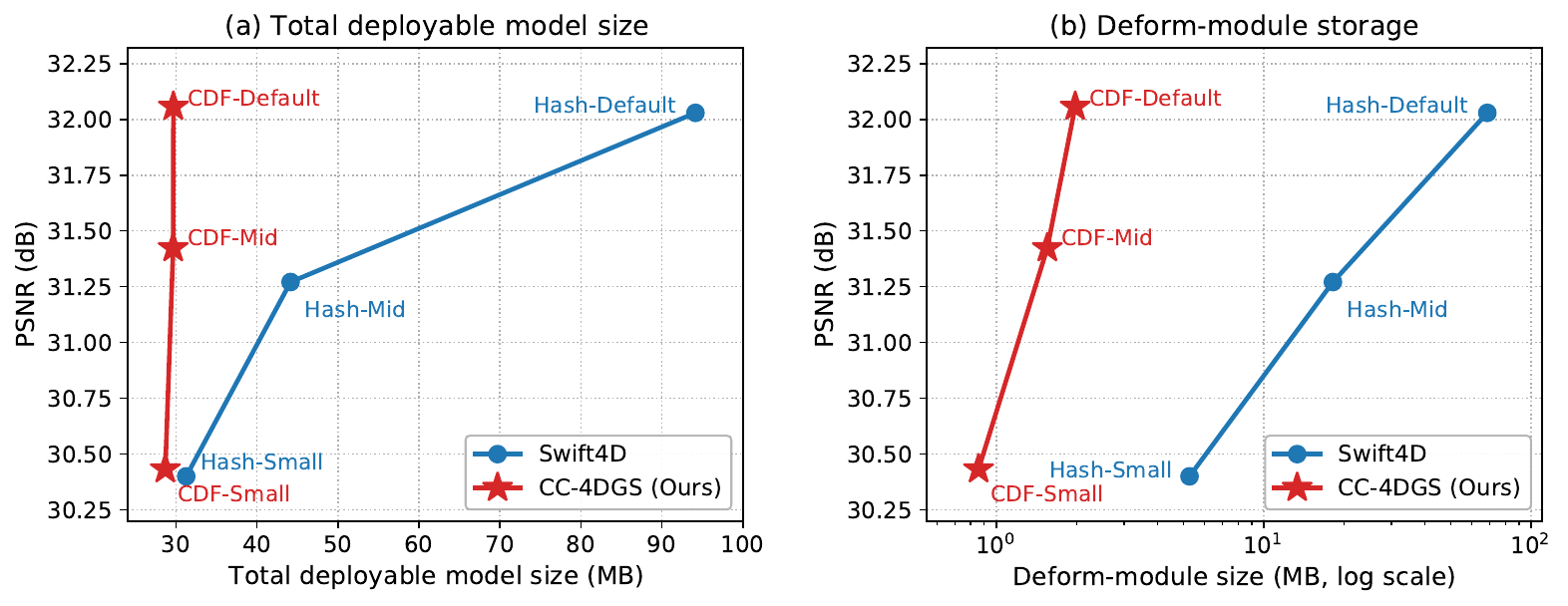}
\caption{Rate--distortion operating points on N3DV, averaged over six scenes. We report three operating points per method under a fixed CCA profile (CCA-Default). For CC-4DGS, we vary the CDF capacity (CDF-Small / Mid / Default); for Swift4D, we vary the hash-grid capacity relative to each scene's default $H^{*}$ (Hash-Small $\equiv H^{*}{-}4$ / Mid $\equiv H^{*}{-}2$ / Default $\equiv H^{*}$). Panel (a) reports the total deployable model size (linear scale); panel (b) isolates the Deform-module storage (log scale). CC-4DGS reaches PSNR comparable to the default Swift4D configuration while requiring substantially smaller total and Deform-module storage.}
\label{fig:rd_neu3d}
\end{figure}

Table~\ref{tab:neu3dstorage} and Table~\ref{tab:neu3dstorage_breakdown} decompose the default storage budget into point-cloud and deformation components. To evaluate whether this storage advantage persists beyond the default operating point, Fig.~\ref{fig:rd_neu3d} reports rate--distortion operating points obtained by down-sweeping each method's deformation capacity from its scene-specific default, averaged over six N3DV scenes. The CCA profile is fixed at CCA-Default for both methods. For CC-4DGS we vary the CDF capacity (CDF-Small / Mid / Default); for Swift4D we vary the hash-grid capacity relative to each scene's default hash-grid size $H^{*}$ (Hash-Small $\equiv H^{*}{-}4$ / Mid $\equiv H^{*}{-}2$ / Default $\equiv H^{*}$). Panel~(a) reports total deployable model size, and panel~(b) isolates the Deform-module storage. Both methods span a similar PSNR range across the sweep ($\approx 30.4$--$32.0$~dB), but their storage envelopes differ by an order of magnitude: all three CC-4DGS configurations stay under $30$~MB total and under $2$~MB in Deform-module storage, whereas Swift4D spans $31$--$94$~MB total and $5$--$68$~MB in the Deform module across the same capacity tiers. Panel~(b) makes this gap explicit: even Swift4D's most aggressively shrunk hash-grid configuration uses more Deform-module storage than CC-4DGS at its full capacity, confirming that the storage advantage is not merely due to point-cloud compression but also reflects the compact CDF representation.
\begin{table}[h]
\centering
\caption{Storage breakdown of the Deform module on the N3DV dataset. \textit{Deform-Grid} denotes the multi-resolution grids or hash tables used for deformation, and \textit{Deform-Net} refers to the weights of the deformation MLP. The \textbf{Best} results for each column are highlighted.}
\label{tab:neu3dstorage_breakdown}
\setlength{\tabcolsep}{3pt}
\begin{tabular}{l
                S[table-format=3.1]
                S[table-format=3.1]
                S[table-format=3.1]}
\toprule
\textbf{Model} &
\multicolumn{1}{c}{\makecell{\textbf{Size(MB)$\downarrow$}\\\textbf{(Deform-Grid)}}} &
\multicolumn{1}{c}{\makecell{\textbf{Size(MB)$\downarrow$}\\\textbf{(Deform-Net)}}} &
\multicolumn{1}{c}{\makecell{\textbf{Size(MB)$\downarrow$}\\\textbf{(Deform-Total)}}} \\
\midrule
Grid4D~\cite{grid4d}   & 220.9 & \bfseries0.2 & 221.1 \\
Swift4D~\cite{Swift4d} &  67.7 & 0.7 &  68.4 \\
\textbf{Ours}          &   \bfseries0.8 & 1.1 &   \bfseries1.9 \\
\bottomrule
\end{tabular}
\end{table}

Table~\ref{tab:cca_bitrate_sweep} reports a complementary sweep along the CCA bitrate axis. While Fig.~\ref{fig:rd_neu3d} varies the deformation capacity under a fixed CCA profile, here we hold the CDF configuration at its default and vary the CCA profile across \textit{Low}, \textit{Mid}, and \textit{Default} rates. Reducing the CCA profile from \textit{Default} to \textit{Low} shrinks the total model size from $29.67$~MB to $23.46$~MB, a $21\%$ reduction, while PSNR drops by only $0.14$~dB on the 6-scene N3DV average. CCA therefore acts as a controllable bitrate knob that is decoupled from the deformation pathway.

\begin{table}[h]
\centering
\caption{CCA bitrate sweep on the N3DV dataset (6-scene average) under a fixed CDF-Default configuration. The three profiles differ in the per-attribute quantization bit budgets and the residual $K$-means codebook size. ``Scalar attr.'' is the shared bit width applied to scale, opacity, and rotation. Lowering the profile from \textit{Default} to \textit{Low} reduces the total model size by about $21\%$ while losing only about $0.14$~dB PSNR.}
\label{tab:cca_bitrate_sweep}
\setlength{\tabcolsep}{4pt}
\resizebox{\columnwidth}{!}{%
\begin{tabular}{lccccccc}
\toprule
\textbf{Profile} & \makecell{\textbf{AE latent}\\\textbf{(bits)}} & \makecell{\textbf{Codebook}\\\textbf{(entries)}} & \makecell{\textbf{Scalar attr.}\\\textbf{(bits)}} & \makecell{\textbf{Dynamic}\\\textbf{(bits)}} & \makecell{\textbf{Total}\\\textbf{(MB)$\downarrow$}} & \makecell{\textbf{PSNR}\\\textbf{(dB)$\uparrow$}} & \textbf{SSIM$\uparrow$} \\
\midrule
\textit{Low}     & $10$ & $64$  & $8$  & $10$ & $23.46$ & $31.92$ & $0.942$ \\
\textit{Mid}     & $12$ & $128$ & $10$ & $12$ & $26.71$ & $32.04$ & $0.943$ \\
\textit{Default} & $14$ & $256$ & $12$ & $16$ & $29.67$ & $32.06$ & $0.943$ \\
\bottomrule
\end{tabular}%
}
\end{table}

\subsection{Long-sequence Stress Test}
\label{subsec:longsequence}

To probe behavior beyond the standard 300-frame protocol, we run a stress test on the \texttt{flame\_salmon} scene of N3DV at sequence lengths $\{300, 600, 900, 1200\}$ frames. Training iterations and densification/pruning schedules are scaled in proportion to the sequence length so that each frame receives comparable training exposure, and CC-4DGS and Swift4D are compared under identical schedules. As Table~\ref{tab:longsequence} reports, both methods degrade gradually as the sequence becomes longer: CC-4DGS goes from $29.62$~dB at 300 frames to $29.12$~dB at 1200 frames ($-0.50$~dB), while Swift4D goes from $29.48$~dB to $28.24$~dB ($-1.25$~dB), with CC-4DGS retaining up to a $+0.88$~dB advantage at 1200 frames. This single-scene stress test does not prove general long-duration scalability, but indicates that the proposed frequency-based time encoding with depth-3 TimeMLP does not exhibit a disproportionately worse long-sequence failure mode than the 4D hash-grid baseline in this setting.

\begin{table}[h]
\centering
\caption{Long-sequence stress test on the flame\_salmon scene of N3DV. Sequence length is scaled from $300$ to $1200$ frames, with training iterations and densification/pruning schedules scaled in proportion. For each length, the better result between CC-4DGS and Swift4D in each column is shown in \textbf{bold}.}
\label{tab:longsequence}
\setlength{\tabcolsep}{4pt}
\begin{tabular}{lcccc}
\toprule
\textbf{Length} & \makecell{\textbf{CC-4DGS}\\\textbf{PSNR (dB)}$\uparrow$} & \makecell{\textbf{CC-4DGS}\\\textbf{SSIM}$\uparrow$} & \makecell{\textbf{Swift4D}\\\textbf{PSNR (dB)}$\uparrow$} & \makecell{\textbf{Swift4D}\\\textbf{SSIM}$\uparrow$} \\
\midrule
$300$  & $\mathbf{29.62}$ & $0.922$ & $29.48$ & $\mathbf{0.923}$ \\
$600$  & $\mathbf{28.91}$ & $\mathbf{0.917}$ & $28.90$ & $0.915$ \\
$900$  & $\mathbf{29.22}$ & $\mathbf{0.916}$ & $28.61$ & $0.913$ \\
$1200$ & $\mathbf{29.12}$ & $\mathbf{0.915}$ & $28.24$ & $0.908$ \\
\bottomrule
\end{tabular}
\end{table}

%% file: Sections/Ablation_Study.tex
This section evaluates how the key components of CC-4DGS affect reconstruction quality, rendering speed, and memory usage through a series of ablation studies. Long-sequence behavior is evaluated separately as a stress test in Sec.~\ref{subsec:longsequence}; this section focuses on component-level ablations. We first compare single-hash and dense hash encoding to assess their impact on dynamic scene reconstruction (Fig.~\ref{fig:abl_single_dense_hash}). We then analyze per-frame rendering time on six N3DV sequences by selectively enabling vertex reuse and feature caching (Table~\ref{tab:abl_cache_reuse}, Fig.~\ref{fig:abl_grid_zoomin}). Finally, we evaluate the FiLM-based fusion path and the TimeMLP depth and time-encoding format that together define the CDF temporal pathway across all six N3DV scenes (Tables~\ref{tab:film_ablation} and \ref{tab:timemlp_ablation}).
\subsection{Benefits of Dense Multi-Hash Encoding}
\begin{figure}[h]
\centering
\includegraphics[width=1.0\linewidth]{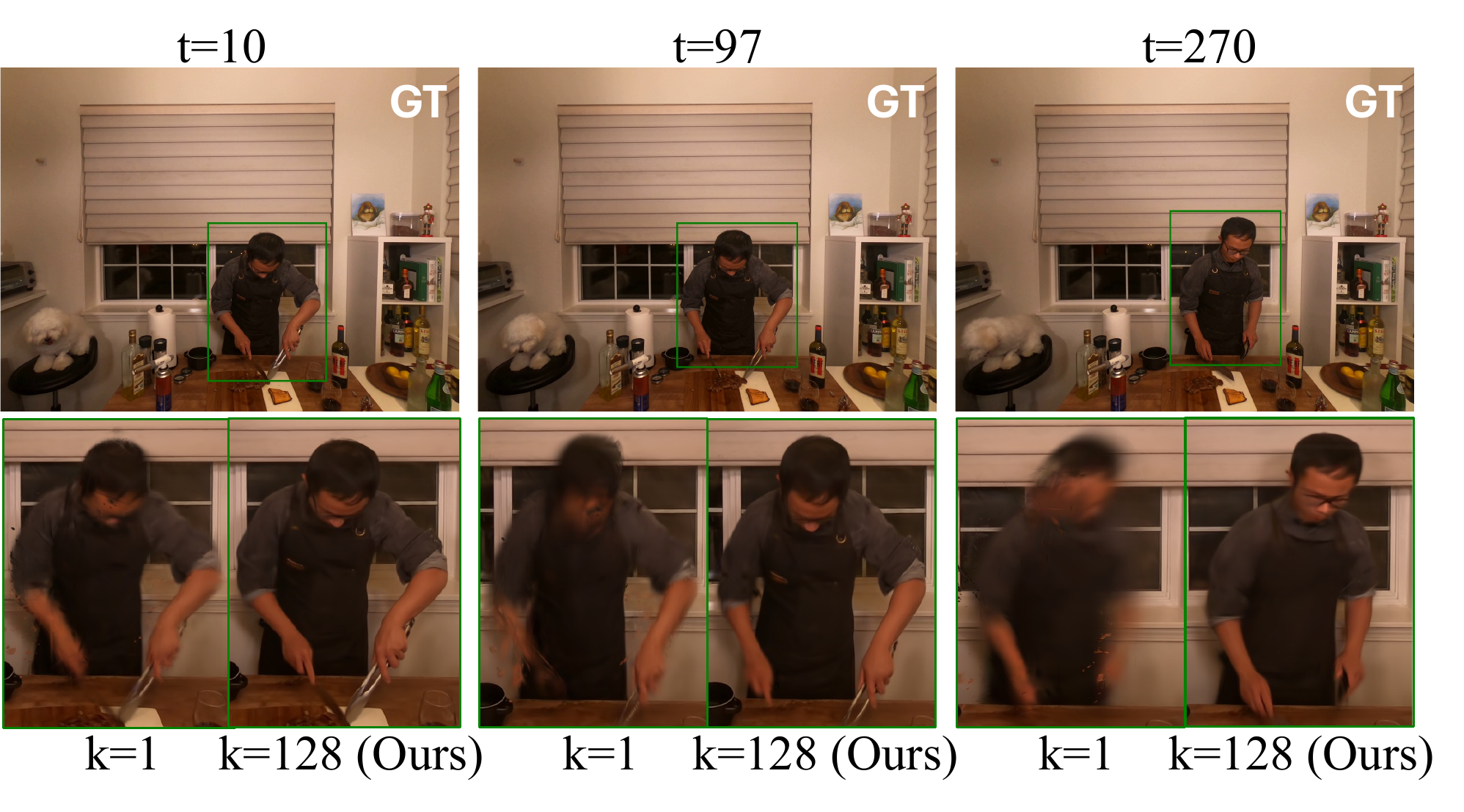}
\caption{Effect of dense hash encoding on dynamic reconstruction quality. Qualitative comparison on the \texttt{cut\_roasted\_beef} scene from N3DV at three timestamps ($t = 10, 97, 270$). For each timestamp, the top row shows the GT. The bottom-left images ($k = 1$) correspond to single-hash encoding, where each Gaussian is mapped using only one hash function. The bottom-right images ($k = 128$, Ours) show the results of dense hash encoding, in which multiple hash functions are aggregated per Gaussian. With $k = 1$, early frames ($t = 10$) appear relatively sharp, but accumulated motion in later frames ($t = 97, 270$) produces severe blur and ghosting around fast-moving regions such as the arms, causing duplicated silhouettes. In contrast, with $k = 128$, these artifacts are greatly suppressed, yielding sharper boundaries and more consistent temporal reconstruction. As deformation behavior is unaffected by CCA, this ablation is evaluated on the raw CDF output without CCA compression.
}
\label{fig:abl_single_dense_hash}
\end{figure}

This experiment evaluates the effect of dense hash encoding by varying the number of hash functions per Gaussian ($k$). For a fair comparison, the voxel grid resolution and bucket size are fixed, and only the single-hash ($k=1$) and dense ($k=128$) configurations are compared. With $k=1$, different Gaussians are more likely to map to the same hash index, leading to blur and ghosting artifacts in regions of fast motion or fine texture. As illustrated in Fig.~\ref{fig:abl_single_dense_hash}, the \texttt{cut\_roasted\_beef} scene shows accumulating motion blur under $k=1$, whereas $k=128$ preserves clear silhouettes and consistent poses over time.
The dense configuration combines multiple hash outputs to generate richer input features, reducing the impact of hash collisions and enabling finer spatiotemporal detail to be captured at the same voxel resolution. This results in a PSNR improvement from 31.37 dB to 33.38 dB, with the largest gains observed in later frames. These findings indicate that dense hash encoding effectively mitigates collision-induced degradation and allows stable reconstruction quality across time, even when using a limited-resolution voxel grid.

\subsection{Acceleration Effects of Vertex Reuse and Feature Caching}
\label{subsec:effect_featurecaching}
\begin{figure}[t]
\centering
\includegraphics[width=1.0\linewidth]{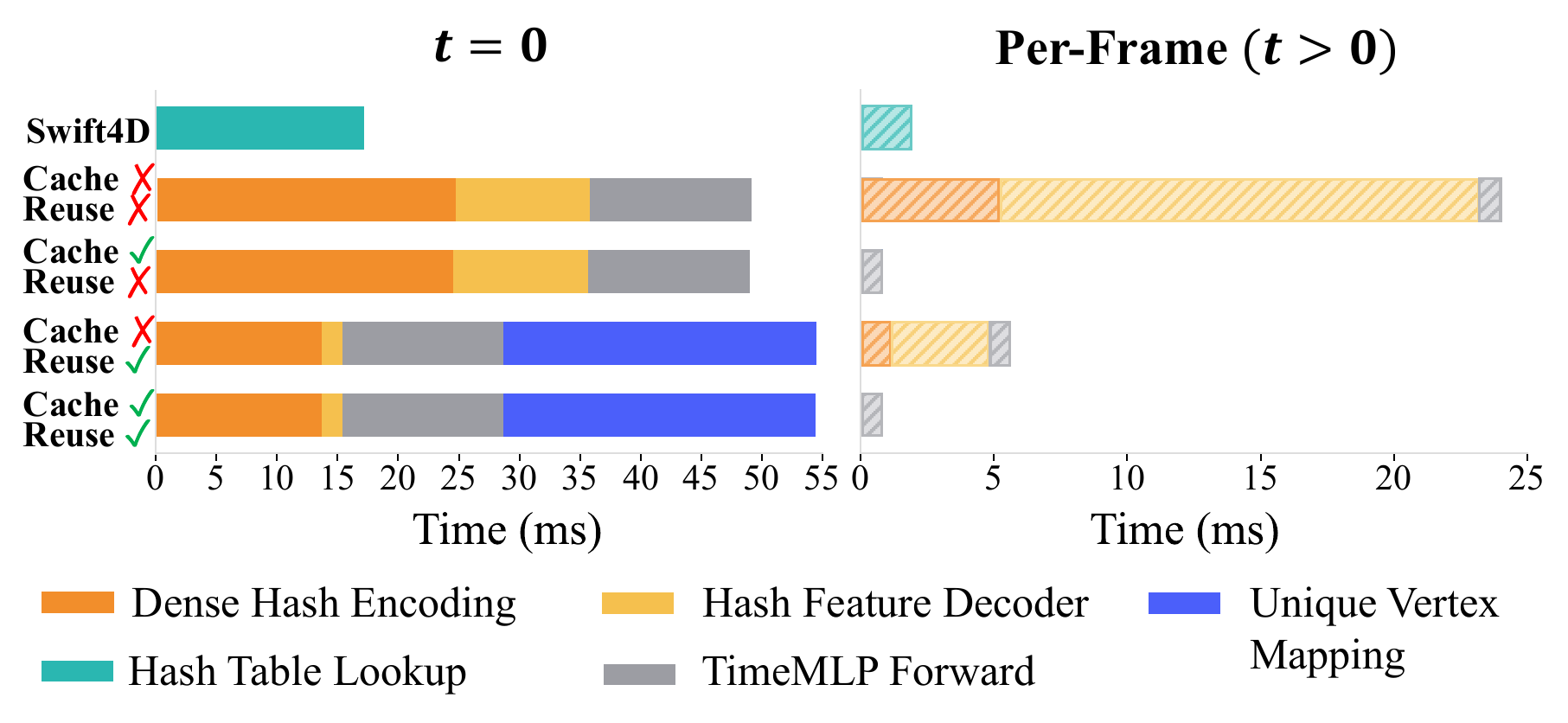}
\caption{Breakdown of the hash-grid path (``grid'' in Table~\ref{tab:abl_cache_reuse}) at the first frame ($t=0$, left) and in the steady state ($t>0$, right) on the N3DV dataset.
Bars correspond to Swift4D and our four variants, where ``Cache'' denotes feature caching and ``Reuse'' denotes vertex reuse.
The stacked segments decompose the grid time into dense hash encoding, deformation hash MLP forward, hash-table lookup or time-MLP forward, and the unique-vertex mapping stage that is present only when vertex reuse is enabled.}
\label{fig:abl_grid_zoomin}
\end{figure}

We evaluate the impact of vertex reuse and feature caching on rendering speed and peak GPU memory. Since CC-4DGS separates spatial and temporal feature computation, enabling feature caching allows spatial hash features to be computed once and reused across all subsequent frames, requiring only the time-dependent branch to be evaluated. Vertex reuse further reduces redundant hash encodings by merging duplicated spatial vertices into a unique set, decreasing both computation and memory usage.
Table~\ref{tab:abl_cache_reuse} and Fig.~\ref{fig:abl_grid_zoomin} break down rendering time into the “grid’’ path (dense hash encoding and decoder operations) and the “other’’ path (deformation network, rasterization, and shading). When caching is enabled, the grid time in the steady state ($t>0$) drops to approximately 0.8\,ms, which is faster than the 1.8\,ms required by Swift4D’s hash-table lookup. This improvement arises because Swift4D must repeatedly fetch features from large hash tables and perform interpolation at every frame, whereas CC-4DGS reuses precomputed spatial features.
Despite this advantage, the end-to-end FPS of Swift4D remains slightly higher due to its highly optimized CUDA kernels (tiny-cuda-nn), while CC-4DGS relies on standard PyTorch operations. Thus, the remaining performance gap stems from implementation-level differences rather than limitations of the CC-4DGS architecture.
Vertex reuse yields additional benefits by amortizing the one-time cost of building the unique-vertex map. Although $t=0$ incurs extra preprocessing overhead, the reduced number of active grid vertices in later frames significantly lowers cumulative rendering time and peak memory usage. This explains why the full configuration achieves the lowest peak memory and why reuse-only notably improves the total runtime compared to the no-cache, no-reuse baseline.
Finally, full and cache-only configurations show nearly identical steady-state grid time (0.8\,ms), and the small difference in total sequence time (about 65\,ms) is within the range of rasterization/other-module variation and measurement noise. This indicates that feature caching alone nearly saturates grid-path efficiency, while vertex reuse mainly contributes to memory reduction and long-term runtime savings.

\begin{table}[t]
\centering
\caption{Rendering time and peak GPU memory on the N3DV dataset (6 scenes).  Time measurements are in milliseconds (ms).  Values for $t=0$ and $t>0$ denote per-frame latency, while \textit{Total} represents the accumulated time for the entire sequence. \textit{Peak mem} is given in gigabytes (GB). 
The \textbf{best} results for each column are highlighted.}
\label{tab:abl_cache_reuse}
\setlength{\tabcolsep}{3pt}
\footnotesize
\begin{tabular}{lcccccc}
\toprule
 & \multicolumn{2}{c}{$t=0$ (ms)} & \multicolumn{2}{c}{$t>0$ (ms)} &
 \textbf{Total}$\downarrow$ & \textbf{Peak mem}$\downarrow$ \\
\cmidrule(lr){2-3} \cmidrule(lr){4-5}
\textbf{Method} & grid & other & grid & other & (ms) & (GB) \\
\midrule
Swift4D~\cite{Swift4d}                & \textbf{17.2} & \textbf{18.1} & 1.8 & \textbf{5.5} & \textbf{2,233} & 5.8 \\
Ours (full: cache+reuse)& 54.5 & 39.6 & \textbf{0.8} & 8.1 & 2,751 & \textbf{5.6} \\
Ours (cache only)       & 49.0 & 41.2 & \textbf{0.8} & 8.3 & 2,816 & 6.1 \\
Ours (reuse only)       & 52.5 & 37.1 & 5.6 & 8.4 & 4,278 & 5.8 \\
Ours (no cache, no reuse)&49.4 & 40.1 & 24.1 & 8.4 & 9,807 & 6.7 \\
\bottomrule
\end{tabular}
\end{table}
\subsection{Effect of FiLM Fusion}
\label{subsec:film_ablation}

The CDF pathway fuses the spatial deformation feature $f_{voxel}$ with the temporal feature $f_{time}$ through FiLM~\cite{film}-based channel-wise modulation rather than direct concatenation, as described in Sec.~\ref{sec:cdf}. To evaluate this design choice on more than a single scene, we report the full FiLM ON/OFF matrix across all six N3DV scenes in Table~\ref{tab:film_ablation}. For each scene we keep the manuscript's default CDF capacity, hash configuration, and training schedule fixed, and only toggle the FiLM module. Evaluations are performed on the raw CDF output without CCA so that quantization noise does not interfere with the fusion comparison.

Across the six scenes the average $\Delta\text{PSNR}{=}\text{ON}{-}\text{OFF}$ is $+1.01$~dB, indicating that FiLM contributes positively on average. The behavior is, however, scene-dependent: FiLM ON improves four scenes (\texttt{coffee\_martini}, \texttt{cook\_spinach}, \texttt{flame\_steak}, \texttt{sear\_steak}) by $+1.90$~dB on average, while FiLM OFF is preferable on two scenes (\texttt{cut\_roasted\_beef}, \texttt{flame\_salmon\_1}) by $-0.79$~dB on average. The Deform-module storage differs by only $2.29$ vs $2.10$~MB between the two settings, so the gap reflects the fusion mechanism rather than a storage-budget asymmetry. We therefore treat FiLM as a CDF design choice that improves reconstruction on average over direct concatenation, with the magnitude varying by scene, rather than a component that is uniformly preferable across every sequence.

\begin{table}[h]
\centering
\caption{FiLM ON/OFF ablation on the six N3DV scenes. For each scene the CDF capacity, hash configuration, and training schedule are held fixed at the manuscript default and only the FiLM module is toggled. $\Delta\text{PSNR}{=}\text{ON}{-}\text{OFF}$. Evaluations use the raw CDF output without CCA compression. The full per-scene matrix is reported so that the scene-dependent behavior of FiLM is visible. In the \emph{Average} row, the better of the two settings per metric is shown in \textbf{bold}.}
\label{tab:film_ablation}
\setlength{\tabcolsep}{3pt}
\resizebox{\columnwidth}{!}{%
\begin{tabular}{lccccc}
\toprule
\textbf{Scene} & \makecell{\textbf{FiLM ON}\\\textbf{PSNR (dB)}$\uparrow$} & \makecell{\textbf{FiLM OFF}\\\textbf{PSNR (dB)}$\uparrow$} & $\Delta$\textbf{PSNR} & \makecell{\textbf{FiLM ON}\\\textbf{SSIM}$\uparrow$} & \makecell{\textbf{FiLM OFF}\\\textbf{SSIM}$\uparrow$} \\
\midrule
\texttt{coffee\_martini}     & $29.33$ & $28.81$ & $+0.52$ & $0.920$ & $0.918$ \\
\texttt{cook\_spinach}       & $32.89$ & $28.70$ & $+4.19$ & $0.951$ & $0.922$ \\
\texttt{cut\_roasted\_beef}  & $32.41$ & $33.42$ & $-1.01$ & $0.949$ & $0.951$ \\
\texttt{flame\_salmon\_1}    & $29.05$ & $29.62$ & $-0.57$ & $0.921$ & $0.922$ \\
\texttt{flame\_steak}        & $33.37$ & $32.67$ & $+0.70$ & $0.958$ & $0.955$ \\
\texttt{sear\_steak}         & $33.82$ & $31.61$ & $+2.21$ & $0.957$ & $0.951$ \\
\midrule
\textbf{Average}             & $\mathbf{31.81}$ & $30.80$ & $+1.01$ & $\mathbf{0.942}$ & $0.936$ \\
\bottomrule
\end{tabular}%
}
\end{table}

\subsection{TimeMLP Depth and Time Encoding Format}
\label{subsec:timemlp_ablation}

The CDF temporal branch encodes the scalar time variable $t$ through frequency-based positional encoding followed by a TimeMLP with three hidden layers (Sec.~\ref{sec:cdf}). To examine both design choices on N3DV, we ablate (i) the TimeMLP depth, by changing the number of hidden layers from three to one or five while keeping frequency-based time encoding, and (ii) the time-encoding format, by replacing the frequency encoding with the raw scalar $t$ while keeping the default depth. All other settings (CDF capacity, hash configuration, training schedule, evaluation iteration) are held at the per-scene manuscript default. Table~\ref{tab:timemlp_ablation} reports the six-scene averages.

Reducing the TimeMLP depth from three to one lowers the average PSNR by $1.01$~dB, while increasing the depth from three to five also reduces PSNR by $0.74$~dB. The default depth therefore sits at the better point in this range under our protocol, with no further gain from a deeper TimeMLP. Replacing the frequency encoding with the raw scalar $t$ at the default depth decreases the average PSNR by $1.66$~dB, indicating that the frequency-based time encoding contributes substantially to learning the deformation field. We therefore retain frequency-based time encoding with a depth-3 TimeMLP as the default CDF temporal pathway.

\begin{table}[h]
\centering
\caption{TimeMLP depth and time-encoding format ablation on the six N3DV scenes. Each variant overrides a single knob (TimeMLP depth or time-encoding format) on top of the per-scene manuscript default; all other settings are preserved. $\Delta\text{PSNR}$ is reported relative to the default. The \textbf{best} result in each column is highlighted.}
\label{tab:timemlp_ablation}
\setlength{\tabcolsep}{4pt}
\begin{tabular}{lccc}
\toprule
\textbf{Variant} & \makecell{\textbf{PSNR (dB)}$\uparrow$} & \textbf{SSIM}$\uparrow$ & $\Delta$\textbf{PSNR (dB)} \\
\midrule
Default (depth $3$ + frequency) & $\mathbf{32.06}$ & $\mathbf{0.943}$ & --- \\
depth $1$ + frequency           & $31.05$ & $0.941$ & $-1.01$ \\
depth $5$ + frequency           & $31.32$ & $0.942$ & $-0.74$ \\
depth $3$ + raw $t$             & $30.40$ & $0.937$ & $-1.66$ \\
\bottomrule
\end{tabular}
\end{table}

\newpage
\subsection{Optimization Cost}
We also report the overall optimization time per scene. On our hardware (a single RTX 4090 GPU), Swift4D converges in about 25 minutes, while 4DGS and Grid4D each take around 60 minutes. Our method finishes in roughly 64--65 minutes per standard 300-frame N3DV scene at the default 20K main-stage budget. Our training time is longer than Swift4D's and comparable to other dynamic 3DGS methods. An optimization time below 1.5 hours per scene is practical for offline content creation, given the storage savings and the preserved real-time rendering speed at test time.
Decomposing this wall-clock gap into schedule length and per-iteration cost, the measured main-stage per-iteration time is about $90$~ms for Swift4D and about $169$~ms for CC-4DGS, reflecting an approximately $1.9\times$ per-iteration computation overhead. In exchange, the Deform-module storage is reduced from $68.4$~MB in Swift4D to $1.9$~MB in CC-4DGS (Table~\ref{tab:neu3dstorage_breakdown}); that is, CC-4DGS trades additional optimization computation for substantially smaller deployable deformation storage. We note that this per-iteration overhead is largely implementation-level rather than intrinsic to the CDF: the decoder is currently built from standard PyTorch operators (such as LayerNorm and Mish) rather than a fused CUDA kernel, and re-expressing it with fused-kernel-compatible operations so that it can run on optimized fused-MLP backends can substantially reduce this cost.

%% file: Sections/Limitations.tex
As Sec.~\ref{subsec:longsequence} shows, both CC-4DGS and Swift4D exhibit gradual quality decrease as the sequence length grows. This reflects a shared limitation of the canonical-plus-deformation 4DGS family that CC-4DGS belongs to: a fixed-capacity canonical point set together with a fixed-capacity deformation pathway must absorb an increasing temporal extent. Streaming-oriented works such as 4DGV~\cite{4dgv} and 4DGC~\cite{4dgc} address this by partitioning the sequence into group-of-pictures (GoP) units with per-segment models, trading training-time integration for streaming-friendly per-segment storage. These methods represent an alternative direction; hierarchical temporal encoding or GoP-style segmentation within the CC-4DGS framework is left as future work.

The computation-over-storage principle behind CC-4DGS, in which a stored trainable hash table is replaced by a small neural decoder evaluated on the fly, has natural extensions to other dynamic 4DGS methods that use a trainable deformation hash grid, such as Swift4D~\cite{Swift4d}, Grid4D~\cite{grid4d}, and 4DGV~\cite{4dgv}. Static representations such as Instant-NGP~\cite{ingp} and hash-grid compression schemes such as HAC~\cite{chen2024hac} differ in rendering-query count, entropy and rate-distortion objectives, and per-vertex compression baselines, and would require a separate design rather than a direct port.

CC-4DGS adopts the dynamic-score-based decomposition of Swift4D~\cite{Swift4d} to gate the deformation pathway, while the dynamic mask itself is learned from scratch. Recent dynamic-static decomposition and streaming methods, including Motion Layering~\cite{4dgv} and Dynamics-Aware Gaussian Splatting Streaming~\cite{dynamics_aware_streaming}, estimate dynamic and static regions either as a preprocessing step or jointly with reconstruction. These external motion or dynamic-region priors could replace or initialize the dynamic mask in CC-4DGS, potentially shortening the dynamic-score warm-up stage. Integrating such priors is a complementary direction and is left as future work.